\documentclass[pdflatex,sn-nature]{sn-jnl}

\usepackage{multirow}%
\usepackage{amsmath,amssymb,amsfonts}%
\usepackage{amsthm}%
\usepackage{mathrsfs}%
\usepackage[title]{appendix}%
\usepackage{textcomp}%
\usepackage{newunicodechar}
\newunicodechar{−}{\textminus}
\usepackage{manyfoot}%
\usepackage{algorithm}%
\usepackage{algorithmicx}%
\usepackage{algpseudocode}%
\usepackage{listings}%
\usepackage[most]{tcolorbox} 
\usepackage{multicol}   
\newcounter{mybox}
\usepackage{tabularx}
\usepackage{booktabs}
\usepackage{colortbl}
\usepackage{caption}
\usepackage{graphicx}
\usepackage{subcaption}
\usepackage{capt-of}
\usepackage{makecell}
\usepackage[section]{placeins} 
\usepackage{hyperref}
\definecolor{headerbg}{RGB}{250, 225, 205}
\usepackage[switch]{lineno} 
\newcolumntype{L}{>{\raggedright\arraybackslash}X}
\newtcolorbox{mhbox}[1][]{
  breakable,
  enhanced,
  colback=gray!12,      
  colframe=gray!12,     
  boxrule=0pt,
  left=1em, right=1em, top=0.6em, bottom=0.9em,
  before skip=10pt, after skip=10pt,
  overlay={
    \draw[line width=0.6pt, black]
      ([xshift=1em,yshift=-2.2em]frame.north west) -- ([xshift=-1em,yshift=-2.2em]frame.north east);
  },
  #1
}

\theoremstyle{thmstyleone}%
\theoremstyle{thmstyletwo}%

\theoremstyle{thmstylethree}%
\newcounter{snote}

\newcounter{edtab}

\DeclareCaptionLabelSeparator{pipe}{\ \textbar\ }
\makeatletter
\newcommand{\fakelabel}[2]{%
  \protected@write\@auxout{}{\string\newlabel{#1}{{#2}{\thepage}}}%
}
\makeatother

\begin{document}


\title{Large language models exhibit unreliable updating of clinical judgment as patient evidence evolves}

\author{\fnm{Min} \sur{Zeng}}
\author*{\fnm{Rui} \sur{Zhang}}\email{ruizhang@umn.edu}

\affil{\orgdiv{Division of Computational Health Sciences, Department of Surgery}, \orgname{University of Minnesota}, \orgaddress{\city{Minneapolis}, \postcode{55455}, \state{MN}, \country{USA}}}

\abstract{
Large language models (LLMs) are increasingly explored for clinical reasoning, but whether they appropriately revise judgments as patient evidence evolves remains unclear. We evaluated longitudinal belief updating using matched intensive-care trajectories from electronic health records. Across diverse LLMs, conditioning on a preceding judgment more often increased than reduced prediction error when estimates changed, replicated for a second endpoint. Controlled interventions revealed two failure modes. First, with preceding assessment fixed, models responded more strongly to worsening than matched improving respiratory evidence; this asymmetry persisted after headroom normalization at moderate and strong evidence levels. Second, with current evidence fixed, increasing prior risk from 10\% to 90\% shifted estimates by 26.2 percentage points, demonstrating causal influence of prior model beliefs. Prompting did not restore reliable updating. Evidence-Validated Longitudinal Update (EVLU) identified fewer, more reliable revisions, revealing a reliability–coverage trade-off. These findings establish longitudinal belief updating as a distinct dimension of LLM reliability.
}

\maketitle

\section*{Introduction}\label{sec1}

Large language models (LLMs) have demonstrated strong capabilities across
medical question answering, clinical knowledge assessment, diagnostic
reasoning, summarization, and decision support
\cite{singhal2023clinical,thirunavukarasu2023large,vanveen2024adapted,
singhal2025expert}, motivating increasing interest in their integration into
clinical workflows. However, much of the supporting evidence still comes
from evaluations that treat model outputs as isolated decisions. A systematic review of
healthcare LLM evaluations found that only a small minority used real
patient-care data and that accuracy remained the dominant evaluation
dimension \cite{bedi2025testing}. Such evaluations characterize what a
model concludes at a particular point in time, but clinical care is
inherently longitudinal: patient physiology evolves, new evidence
accumulates, and assessments that were appropriate at one time point may
need to be maintained, strengthened, or revised as the clinical state
changes.

Recent work has moved beyond static medical question answering toward
sequential clinical evaluation. Clinician-generated EHR tasks, simulated
clinical interactions, and stepwise diagnostic workflows have revealed
limitations not apparent from examination-style accuracy alone
\cite{fleming2024medalign,johri2025evaluation,hager2024evaluation,
rao2026large}, while performance can vary with the quantity or order of
clinical information \cite{hager2024evaluation}. Studies of longitudinal
electronic health records further suggest that longer clinical histories do
not necessarily yield reliable temporal reasoning
\cite{fleming2024medalign,kruse2025large,cui2025timer}. Thus, strong performance on static or aggregate clinical benchmarks does not establish whether models can appropriately revise clinical judgments as patient evidence evolves.

Longitudinal use introduces an additional challenge: once a model has
generated a clinical assessment, that assessment may itself become part of
the context for subsequent reasoning. This can provide useful continuity, but also creates a pathway through which previous model assessments may influence later judgments beyond what current patient evidence supports.
More generally, language models have been shown to struggle with belief
revision when new information requires modification of an earlier conclusion
\cite{wilie2024belief}, to exhibit sensitivity to prior user positions and
contextual framing \cite{sharma2023sycophancy,mahajan2025cognitive,
sheppert2026anchoring}, and, in conflicting-context settings, to preferentially
rely on model-generated context even when external context is more reliable
\cite{tan2024generatedcontext}. Multi-turn evaluations have further identified
susceptibility to error propagation across conversational turns
\cite{kwan2024mteval}. An appropriate longitudinal clinical reasoner should preserve a preceding judgment when it remains supported, revise it when contradicted by new evidence, and adjust the magnitude of revision according to the strength of that evidence.
Failure to do so could allow inaccurate assessments to propagate across
time or produce systematically different responses to clinically
comparable evidence of deterioration and recovery.

Existing evaluations do not disentangle evolving patient evidence from the model's preceding judgment because both change together during clinical trajectories. Complementary controlled interventions are therefore needed to independently manipulate each source of longitudinal context: manipulating newly presented clinical evidence while holding the preceding model judgment fixed, and independently manipulating the preceding judgment while holding the current clinical evidence unchanged. Such a design can distinguish evidence-responsive updating from dependence induced by model-generated history and identify longitudinal updating patterns that are invisible to conventional snapshot-level evaluation.

Here, we systematically characterize longitudinal clinical belief updating in LLMs using matched intensive-care trajectories derived from electronic health records. We use “belief” operationally to denote the model’s expressed clinical assessment or risk estimate, without implying human-like mental states. We compare independent assessments with longitudinal assessments conditioned on the model's preceding judgment across diverse general-purpose and medically adapted LLMs, with cross-endpoint replication for vasopressor initiation. Paired controlled interventions separately manipulate evolving respiratory evidence and the supplied preceding model belief, characterizing directional evidence responsiveness and causally isolating prior-belief influence. Finally, we test evidence-first and anti-anchoring prompting and introduce an Evidence-Validated Longitudinal Update (EVLU) strategy that retains context-induced revisions only when corroborated by evolving objective respiratory evidence. Together, these experiments identify longitudinal belief updating as a distinct dimension of LLM reliability: model-generated beliefs and evolving patient evidence are not interchangeable forms of context.

\section*{Results}\label{sec2}

\subsection*{Construction of a matched longitudinal clinical cohort}\label{sec:2.1}
\begin{figure*}[t]
    \centering
    \includegraphics[width=\textwidth]
    {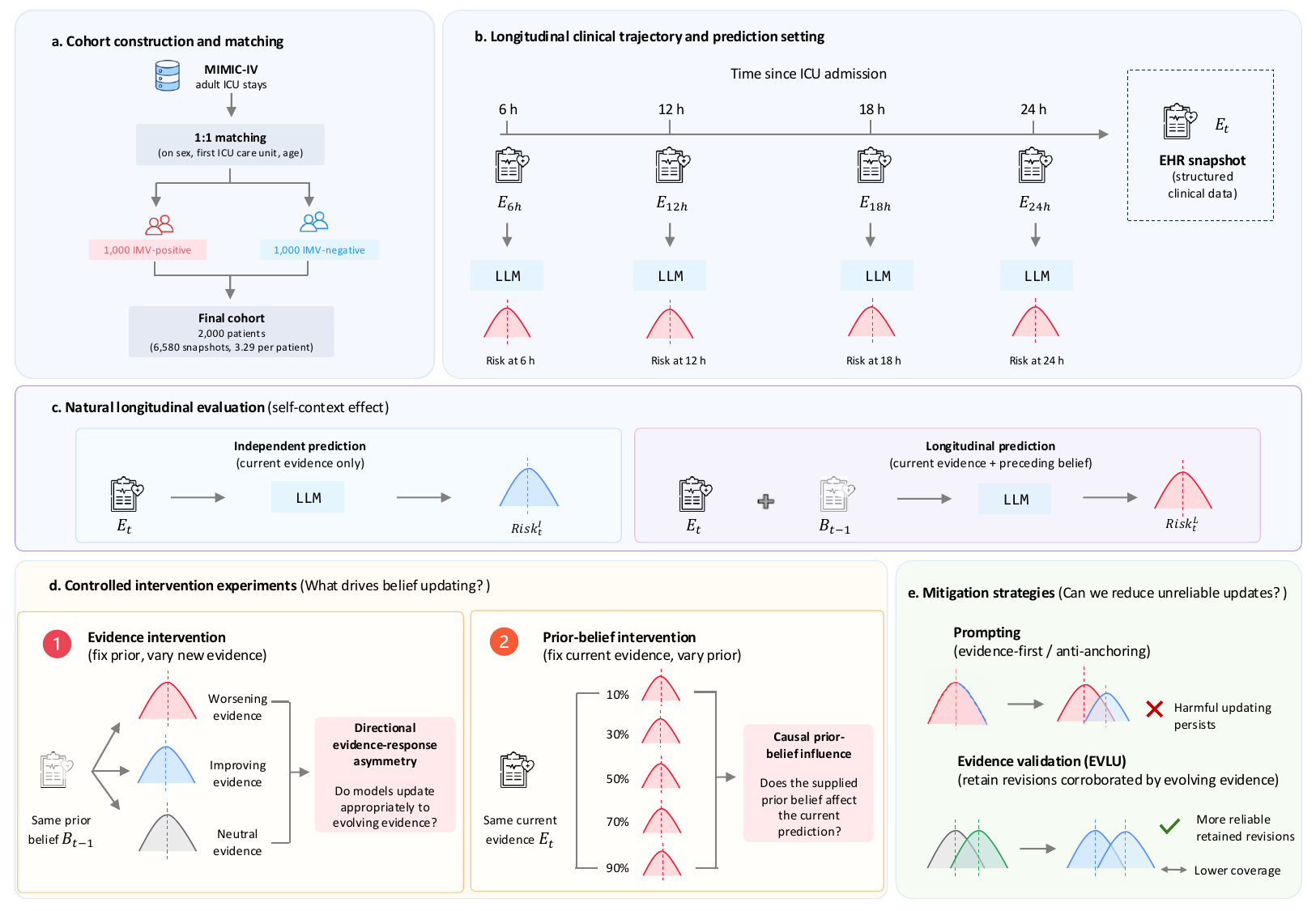}
    \vspace{1mm}
    \caption{
    \textbf{Study design for evaluating longitudinal clinical belief updating as patient evidence evolves.}
    \textbf{a}, Construction of the matched longitudinal ICU cohort,
    comprising 1,000 ventilation-positive ICU stays and 1,000 matched negative
    ICU stays.
    \textbf{b}, Longitudinal clinical states were represented by sequential
    snapshots after ICU admission, with the model estimating the risk of
    invasive mechanical ventilation within the subsequent 24~h at each
    snapshot.
    \textbf{c}, Natural longitudinal evaluation compared predictions based on
    the current clinical state alone with predictions additionally conditioned
    on the model's preceding assessment.
    \textbf{d}, Controlled interventions independently manipulated evolving
    respiratory evidence and the supplied preceding model belief while holding
    the complementary source of information fixed, enabling characterization
    of directional evidence responsiveness and the causal influence of supplied
    prior beliefs.
    \textbf{e}, Mitigation analyses tested instruction-level prompting and evidence validation; prompting did not eliminate harmful longitudinal updating, whereas EVLU preferentially retained a smaller subset of more reliable revisions at reduced coverage.
    Icon attributions are provided in Table S1.
    }
    \label{fig:study_overview}
\end{figure*}

We constructed a matched longitudinal ICU cohort from
MIMIC-IV~\cite{johnson2023mimic} (Fig.~\ref{fig:study_overview}). The cohort
comprised 1,000 ventilation-positive ICU stays and 1,000 negative stays
matched on sex, first ICU care unit and age. The 2,000 unique patients contributed 6,580 clinical snapshots (mean, 3.29 per stay). All positive-stay snapshots preceded invasive ventilation, preventing incorporation of post-outcome information. This cohort supported natural longitudinal evaluation and controlled evidence interventions.

\subsection*{Longitudinal self-context more often worsens than improves paired prediction accuracy}\label{sec:2.2}


We first examined whether access to the model's preceding assessment altered
subsequent risk estimates under naturally evolving clinical trajectories
(Fig.~\ref{fig:study_overview}c). Across 4,580 post-initial snapshots,
longitudinal predictions were 9.50 percentage points higher than independent
predictions of the same current clinical state on average, with a mean
absolute difference of 14.90 percentage points.

\begin{figure*}[!t]
    \centering
    \begin{minipage}[t]{0.45\textwidth}
        \noindent\textbf{a}\par
        \vspace{0.15em}
        \centering
        \includegraphics[width=\linewidth]
        {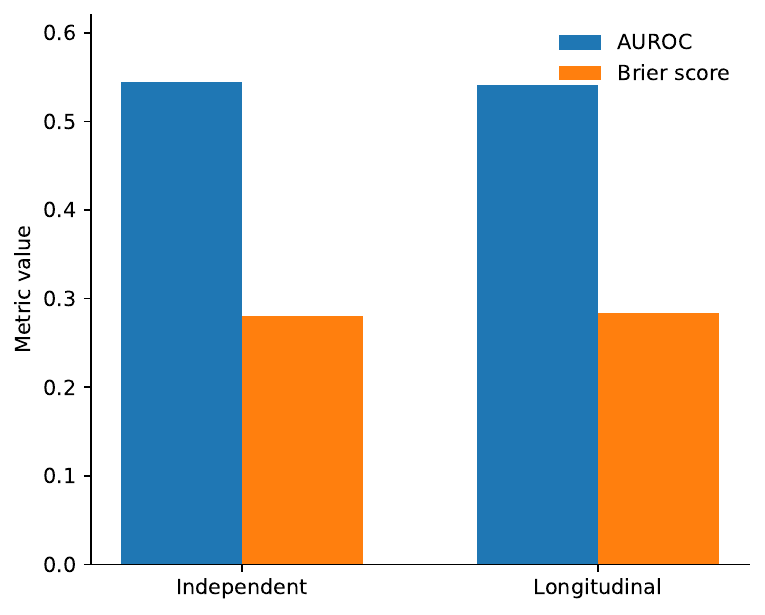}
    \end{minipage}
    \hfill
    \begin{minipage}[t]{0.45\textwidth}
        \noindent\textbf{b}\par
        \vspace{0.15em}
        \centering
        \includegraphics[width=\linewidth]
        {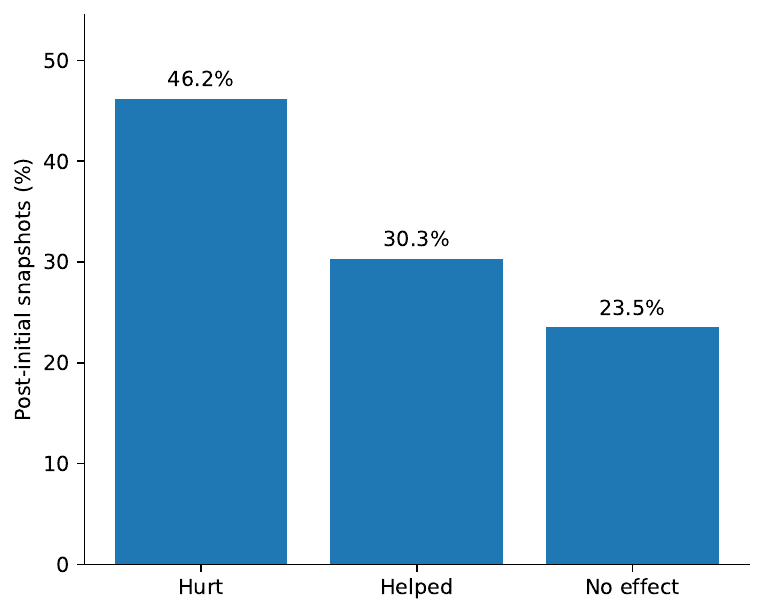}
    \end{minipage}

\vspace{0.5em}
    \begin{minipage}[t]{0.98\textwidth}
        \noindent\textbf{c}\par
        \vspace{0.15em}
        \centering
        \includegraphics[width=\linewidth]
        {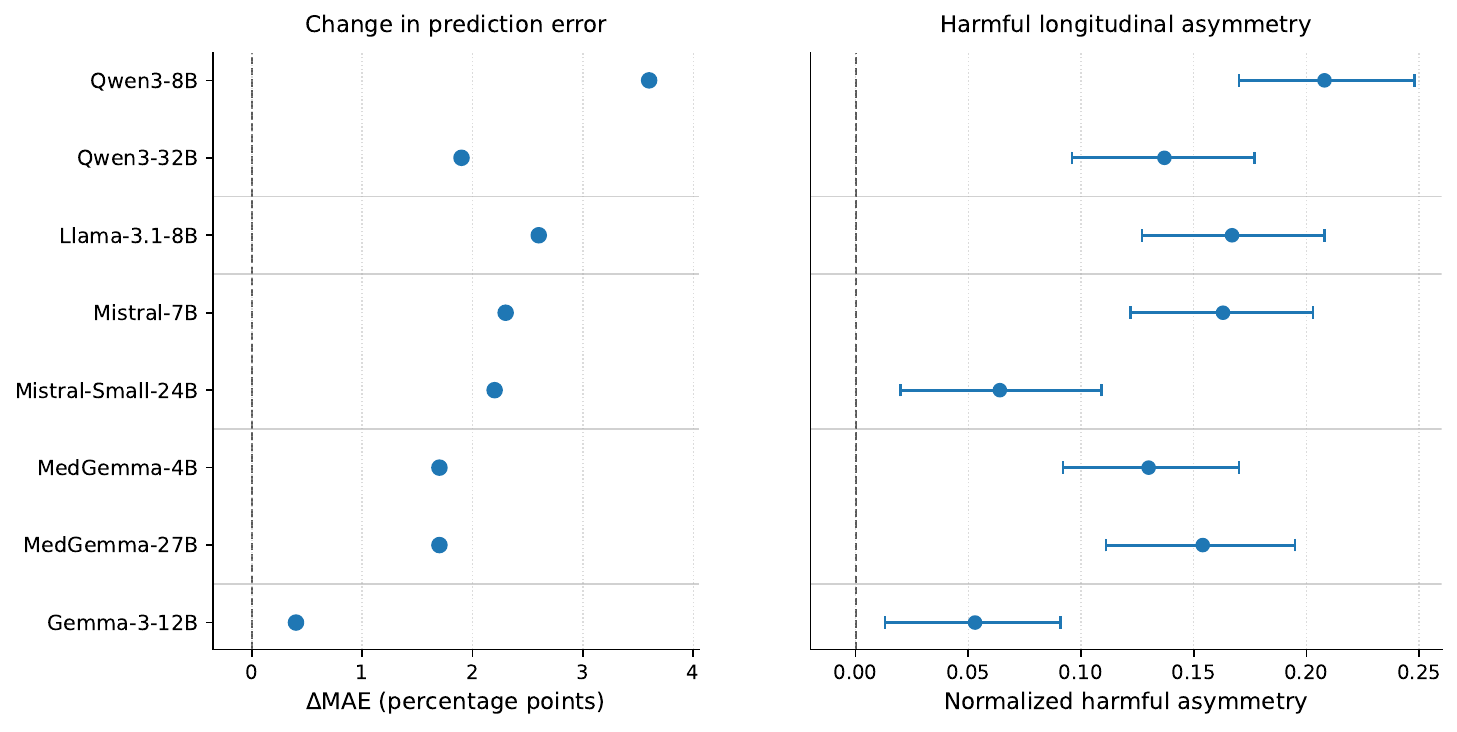}
    \end{minipage}

    \caption{
    \textbf{Longitudinal self-context alters clinical risk estimates and
    produces harmful paired effects across model families.}
    \textbf{a}, Predictive performance under independent and longitudinal
    evaluation for the primary model, measured by AUROC and Brier score
    across 6,580 clinical snapshots.
    \textbf{b}, Effect of longitudinal self-context on absolute prediction
    error across 4,580 post-initial snapshots. Longitudinal context increased
    prediction error more frequently than it reduced error.
    \textbf{c}, Cross-model generalization of harmful longitudinal-context
    effects across eight evaluated models. Left, change in mean absolute
    prediction error under longitudinal relative to independent inference.
    Right, normalized harmful-update asymmetry between harmful and beneficial accuracy-changing effects; positive values indicate a greater frequency of harmful than beneficial effects among accuracy-changing predictions.
    Error bars in the right panel indicate 95\% confidence intervals obtained
    by bootstrap resampling at the ICU-stay level.
    }
    \label{fig:longitudinal_context}
\end{figure*}

These context-induced changes did not improve predictive performance. Across
all 6,580 snapshots, AUROC was 0.545 under independent evaluation and 0.541
under longitudinal evaluation, while Brier score increased from 0.281 to
0.284 (Fig.~\ref{fig:longitudinal_context}a). At the paired-snapshot level,
longitudinal context increased absolute prediction error in 46.20\% of cases,
reduced it in 30.28\%, and left it unchanged in 23.52\%
(Fig.~\ref{fig:longitudinal_context}b), increasing mean absolute error by
3.62 percentage points. Thus, self-context substantially altered subsequent
predictions but was more likely to impair than improve their accuracy.

\subsection*{Harmful longitudinal dependence replicates in hemodynamic deterioration}
\label{sec:2.3}
We next tested whether harmful longitudinal dependence generalized from
respiratory to hemodynamic deterioration. We independently constructed a
second matched cohort for 24-h prediction of continuous vasopressor
initiation, comprising 1,000 vasopressor-positive and 1,000 matched
vasopressor-negative ICU stays. The cohort yielded 6,444 serial assessments
evaluated using the same longitudinal protocol as for invasive ventilation.

The central pattern replicated closely. Across 4,444 post-initial assessments, longitudinal context increased absolute prediction error in 46.7\% of cases, reduced it in 28.8\%, and
left it unchanged in 24.5\%. Mean absolute prediction error increased by
3.44 percentage points (95\% CI, 2.62--4.24), while predicted risk shifted
upward by 10.03 percentage points (95\% CI, 9.33--10.76).
This paired degradation occurred despite a modest increase in aggregate
discrimination (AUROC, 0.574 to 0.582), showing that improved aggregate
ranking can coexist with less reliable longitudinal updating. 
Thus, harmful longitudinal dependence replicated across independently constructed respiratory and hemodynamic deterioration cohorts.

\subsection*{Harmful longitudinal dependence generalizes across model families}
\label{sec:2.4}
We tested whether harmful longitudinal dependence generalized beyond the primary Qwen3-8B model by repeating the independent-versus-longitudinal evaluation across eight general-purpose and medically adapted models spanning the Qwen, Llama, Mistral, MedGemma, and Gemma families (4B--32B parameters). All models used the same 2,000-patient cohort, clinical timelines, prediction target, and inference protocol.

Mean absolute prediction error increased by 0.4--3.6 percentage points under longitudinal relative to independent inference in all eight models (Fig.~\ref{fig:longitudinal_context}c; Table S2). Aggregate discrimination was less consistent: AUROC decreased in six models but increased slightly in MedGemma-27B-Text-IT and Gemma-3-12B-IT. Thus, aggregate ranking performance did not reliably track effects on paired prediction accuracy.

Among post-initial snapshots whose prediction accuracy changed, the harmful fraction exceeded 50\% in every model (52.63--60.41\%). Normalized harmful-update asymmetry was correspondingly positive across all eight models (0.053--0.208), with 95\% confidence intervals excluding zero in every case (Fig.~\ref{fig:longitudinal_context}c; Table S3).

Normalized harmful-update asymmetry decreased from 0.208 to 0.137 between Qwen3-8B and Qwen3-32B and from 0.163 to 0.064 between the two evaluated Mistral scales, but increased from 0.130 to 0.154 between MedGemma-4B-IT and MedGemma-27B-Text-IT. Thus, although effect magnitude varied across model families and scales, larger models did not consistently attenuate harmful longitudinal dependence.

\subsection*{Belief revision is directionally appropriate and dose-responsive to controlled evolving evidence}
\label{sec:2.5}

We next tested whether the model revised its beliefs appropriately in
response to controlled changes in respiratory evidence
(Fig.~\ref{fig:study_overview}d). Of 4,580 real longitudinal transitions,
4,545 contained the respiratory measurements required for intervention
construction. For each eligible transition, we held the preceding context
and prior model assessment fixed while introducing worsening or improving
respiratory evidence at subtle, moderate, or strong intensities.

\begin{figure*}[!htbp]
\centering

\begin{minipage}[t]{0.48\textwidth}
    \noindent\textbf{a}\par
    \vspace{0.15em}
    \centering
    \includegraphics[width=\linewidth]
    {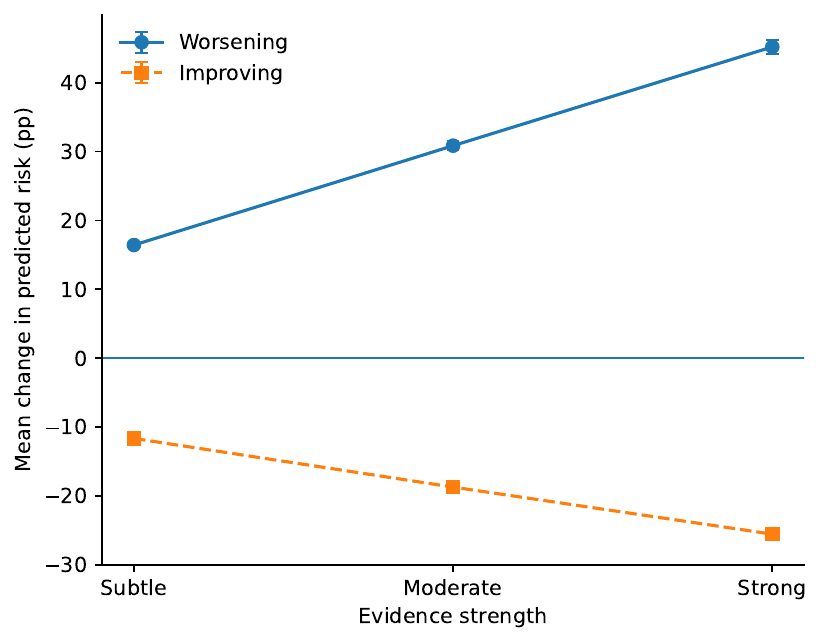}
\end{minipage}
\hfill
\begin{minipage}[t]{0.48\textwidth}
    \noindent\textbf{b}\par
    \vspace{0.15em}
    \centering
    \includegraphics[width=\linewidth]
    {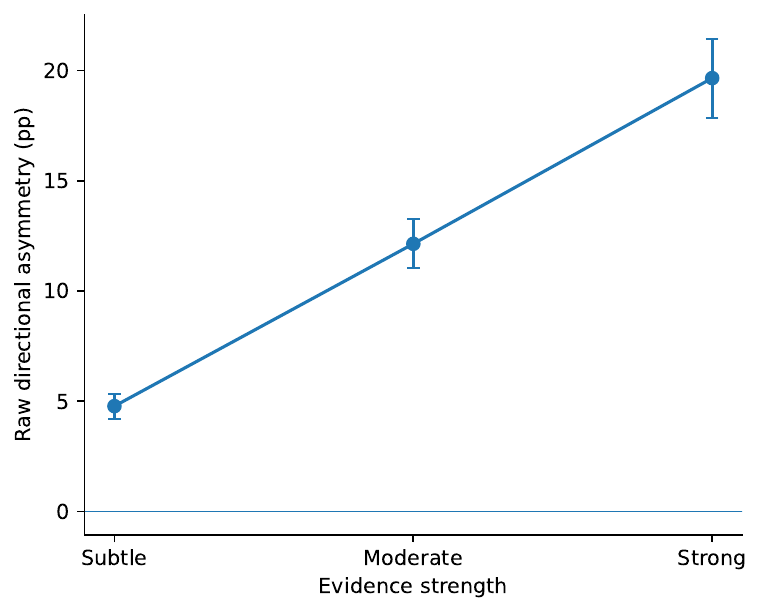}
\end{minipage}

\vspace{0.5em}

\begin{minipage}[t]{0.82\textwidth}
    \noindent\textbf{c}\par
    \vspace{0.15em}
    \centering
    \includegraphics[width=\linewidth]
    {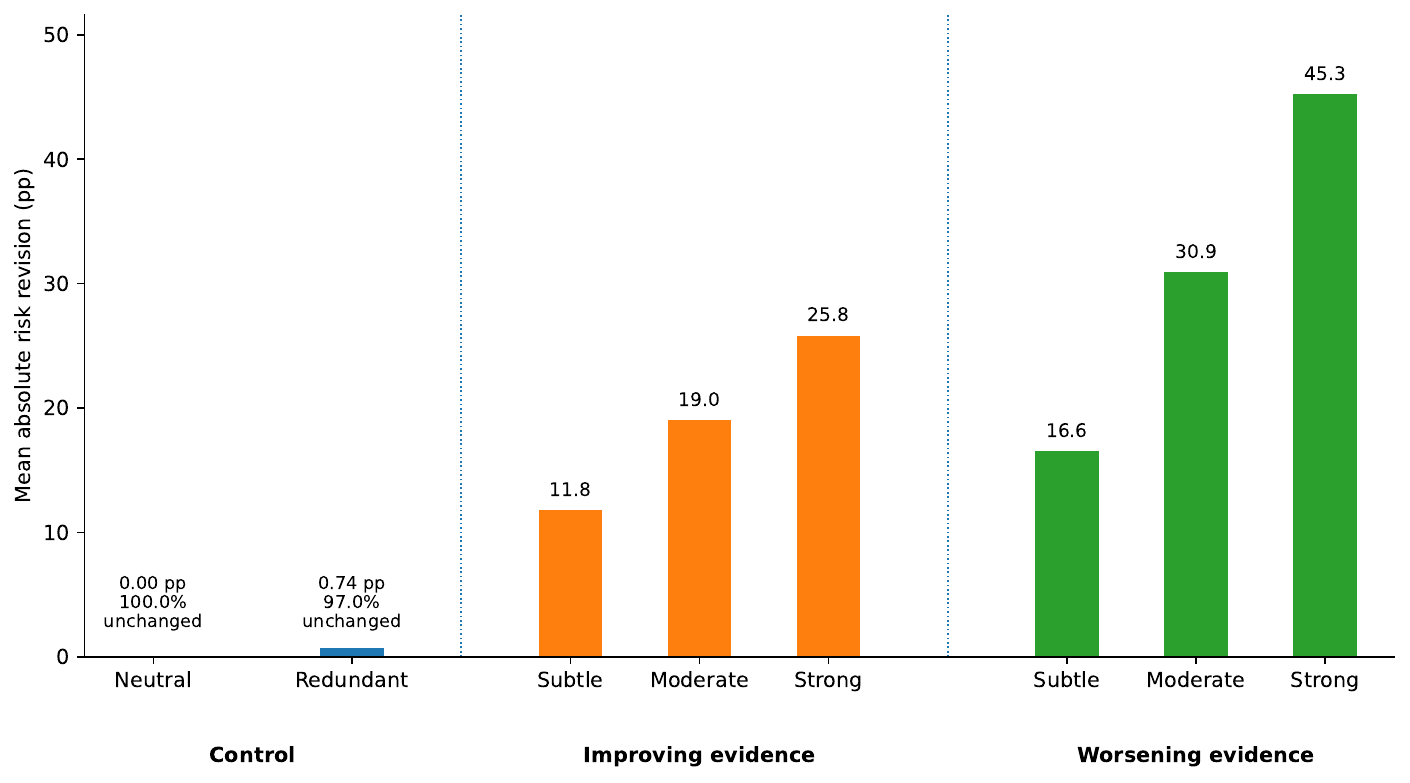}
\end{minipage}

\caption{
\textbf{Controlled respiratory evidence produces selective, dose-dependent,
and directionally asymmetric belief revision.}
\textbf{a}, Mean change in predicted 24-h invasive ventilation risk following
worsening and improving evidence at subtle, moderate, and strong intervention
strengths.
\textbf{b}, Raw directional asymmetry across evidence strengths. Asymmetry
was defined as the excess magnitude of deterioration-directed relative to
improvement-directed belief revision; positive values indicate stronger
updating in response to worsening evidence.
\textbf{c}, Mean absolute changes in predicted ventilation risk under neutral
and redundant control interventions compared with clinically informative
improving and worsening interventions. Neutral and redundant evidence elicited little revision on average.
Error bars in \textbf{a}--\textbf{c} indicate 95\% confidence intervals obtained by bootstrap resampling at the ICU-stay level.
}
\label{fig:counterfactual}
\end{figure*}

Worsening evidence increased mean predicted risk by 16.43, 30.86, and 45.21 percentage points under subtle, moderate, and strong perturbations, respectively, with directionally appropriate increases in 97.32--99.85\% of transitions (Fig.~\ref{fig:counterfactual}a). Improving evidence decreased mean predicted risk by 11.66, 18.72, and 25.55 percentage points, respectively, with directionally appropriate decreases in 86.84--91.90\% of transitions.

Revision magnitude increased monotonically with evidence strength in
99.93\% of worsening transitions and 97.58\% of improving transitions.
Thus, belief revision tracked both the direction and strength of evolving
evidence, providing a basis for testing whether update magnitude was
symmetric between deterioration and improvement.

\subsection*{Belief updating exhibits strength-dependent directional asymmetry}
\label{sec:2.6}

On the raw probability scale, worsening evidence produced larger
belief revisions than matched improving evidence at all three intervention
strengths. Mean raw asymmetry increased from 4.78 percentage points for subtle evidence
(95\% CI, 4.21--5.34) to 12.14 percentage points for moderate evidence
(95\% CI, 11.02--13.26) and 19.65 percentage points for strong evidence
(95\% CI, 17.86--21.44) (Fig.~\ref{fig:counterfactual}b). Thus, although both directions showed dose-responsive revision, worsening evidence elicited progressively larger updates than matched improving evidence as evidence strength increased.

\subsection*{Directional asymmetry persists after accounting for probability headroom and preceding prediction error}
\label{sec:2.7}

A potential explanation for the raw directional asymmetry is that upward and downward revisions are mechanically constrained by the preceding risk estimate. We therefore normalized each revision by the available probability headroom in its corresponding direction.

\begin{figure*}[!htbp]
\centering

\begin{minipage}[t]{0.48\textwidth}
    \raggedright
    \textbf{a}\\
    \centering
    \includegraphics[width=\linewidth]
    {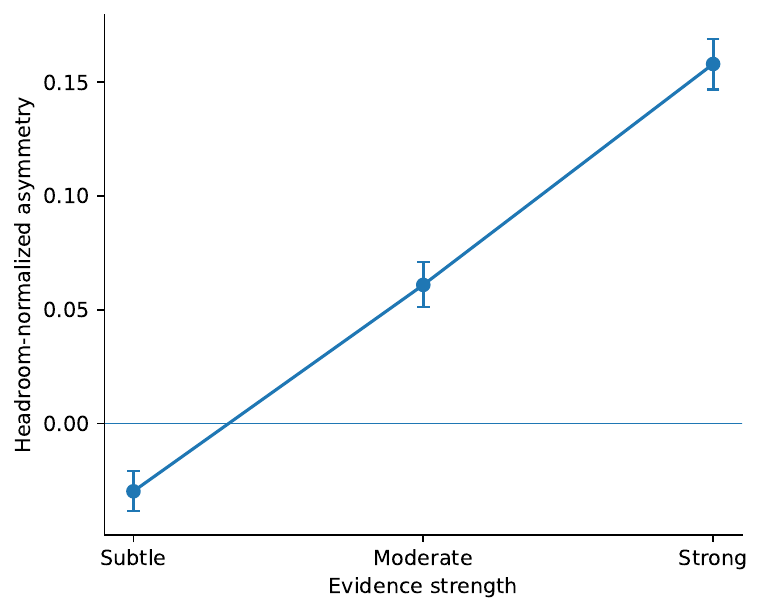}
\end{minipage}
\hfill
\begin{minipage}[t]{0.48\textwidth}
    \raggedright
    \textbf{b}\\
    \centering
    \includegraphics[width=\linewidth]
    {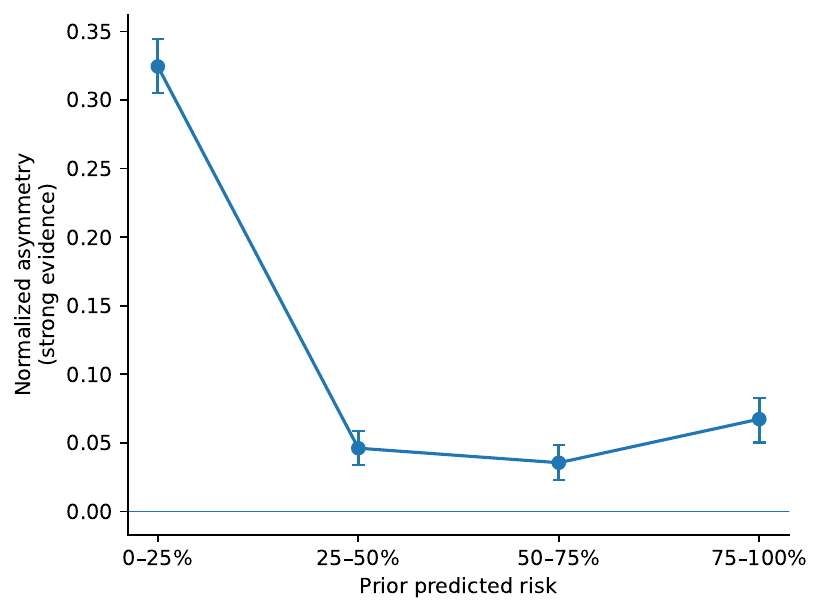}
\end{minipage}

\vspace{0.5em}

\begin{minipage}[t]{0.48\textwidth}
    \raggedright
    \textbf{c}\\
    \centering
    \includegraphics[width=\linewidth]
    {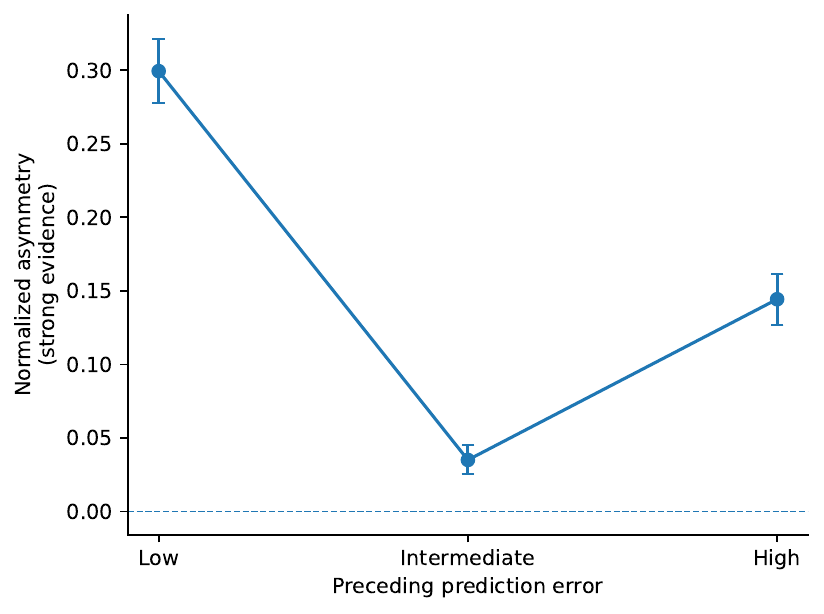}
\end{minipage}

\caption{
\textbf{Directional asymmetry persists after accounting for prior belief headroom and preceding prediction error.}
\textbf{a}, Directional asymmetry after normalizing worsening and improving revisions by the available upward and downward probability ranges, respectively. Moderate and strong evidence retained deterioration-directed asymmetry after normalization.
\textbf{b}, Headroom-normalized asymmetry under strong evidence stratified by the preceding predicted ventilation risk. Positive asymmetry persisted across all four prior-risk strata.
\textbf{c}, Headroom-normalized asymmetry under strong evidence stratified by the absolute error of the preceding risk estimate. Positive asymmetry persisted across low, intermediate, and high prediction-error strata, including transitions with relatively accurate preceding predictions.
Error bars indicate 95\% bootstrap confidence intervals obtained by resampling at the ICU-stay level.
}
\label{fig:headroom}

\end{figure*}

After headroom normalization, subtle evidence showed a small
improvement-directed asymmetry ($-0.030$; 95\% CI, $-0.039$ to $-0.021$),
whereas moderate and strong evidence retained deterioration-directed
asymmetry of 0.061 (95\% CI, 0.051--0.071) and 0.158
(95\% CI, 0.147--0.169), respectively
(Fig.~\ref{fig:headroom}a). Thus, unequal probability headroom contributed to the raw asymmetry under
weak evidence but did not explain the stronger response to moderate and
strong worsening evidence.

Under strong evidence, deterioration-directed asymmetry
remained positive across all four ascending prior-risk strata
(0.324, 0.046, 0.036, and 0.067, respectively;
Fig.~\ref{fig:headroom}b), arguing against a simple ceiling- or floor-effect
explanation.

We next stratified transitions by the absolute error of the preceding risk estimate. Under strong evidence, headroom-normalized asymmetry remained positive across low-, intermediate-, and high-error strata (0.299, 0.035, and 0.144, respectively; Fig.~\ref{fig:headroom}c), including transitions with relatively accurate preceding predictions. Because absolute prediction error is related to both the preceding probability estimate and observed outcome, we additionally stratified by outcome. Deterioration-directed asymmetry remained positive across all six outcome-by-error strata, with 95\% bootstrap confidence intervals excluding zero (Table S4).

The asymmetry was also evident at the individual-transition level. Under
strong evidence, worsening perturbations produced greater proportional
revision than matched improving perturbations in 66.82\% of transitions,
whereas 27.46\% showed the opposite pattern and 5.72\% showed equal
updating. Together, these analyses show that directional asymmetry was not explained by preceding prediction error or outcome composition and was broadly distributed across individual transitions.

\subsection*{Neutral and redundant evidence elicit little belief revision on average}
\label{sec:2.9}

We next tested whether belief revision was specific to clinically
informative evidence rather than a generic response to additional context.
Across the same 4,545 eligible transitions used for the directional
perturbations, neutral evidence produced no change in predicted risk,
whereas redundant evidence produced a mean absolute revision of 
0.74 percentage points and no numerical change in 96.99\% of transitions
(Fig.~\ref{fig:counterfactual}c). In contrast, worsening and improving
evidence elicited substantially larger revisions. Thus, substantial belief updating was not a generic consequence of appending additional context, but occurred predominantly in response to
clinically meaningful changes in respiratory evidence.

\subsection*{Directional asymmetry is consistent across clinical outcome groups}
\label{sec:2.10}

We further examined whether the full strength-dependent pattern of
directional asymmetry was preserved across clinical outcome groups.
Under strong evidence, headroom-normalized asymmetry remained
deterioration-directed in both ventilation-positive and
ventilation-negative trajectories (0.124 and 0.192, respectively).
Moreover, both outcome groups showed a similar strength-dependent transition
from a small improvement-directed asymmetry under subtle evidence to
increasingly deterioration-directed asymmetry as evidence strengthened
(Table S5). Thus, strength-dependent directional asymmetry was not restricted to patients who ultimately required invasive ventilation.

\subsection*{Controlled prior-belief interventions causally shift subsequent risk estimates}
\label{sec:2.11}

Because prior assessments and evolving clinical evidence
co-occur in natural trajectories, these analyses cannot isolate whether the
prior belief itself causally influences the subsequent prediction. We
therefore performed a controlled prior-belief intervention in which the
current clinical evidence was held fixed while the supplied prior
ventilation risk was manipulated to 10\%, 30\%, 50\%, 70\%, or 90\%.

\begin{figure*}[!htbp]
\centering
\includegraphics[width=\textwidth]{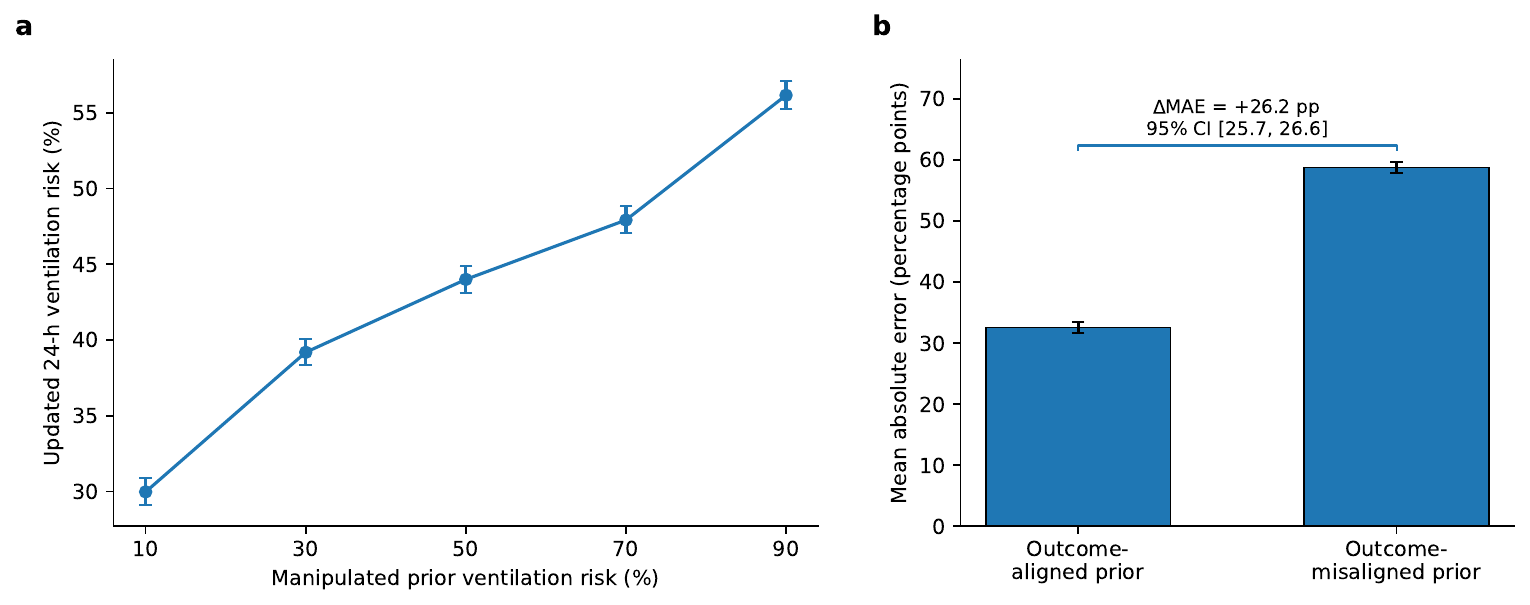}
\caption{
\textbf{Controlled prior-belief interventions causally shift subsequent
clinical risk estimates.}
\textbf{a}, Dose--response relationship between the experimentally
manipulated prior 24-h ventilation risk and the subsequent updated risk
estimate while current clinical evidence was held fixed. Points indicate
mean updated risk estimates and error bars indicate ICU-stay-level bootstrap
95\% confidence intervals.
\textbf{b}, Mean absolute prediction error under outcome-aligned and
outcome-misaligned extreme prior beliefs. For outcome-negative snapshots,
10\% and 90\% priors were treated as aligned and misaligned, respectively;
for outcome-positive snapshots, 90\% and 10\% priors were treated as aligned
and misaligned, respectively. Error bars indicate ICU-stay-level bootstrap
95\% confidence intervals. The annotated difference denotes the increase
in mean absolute error for outcome-misaligned relative to outcome-aligned
priors.
}
\label{fig:prior_intervention}
\end{figure*}

Manipulating the supplied prior produced a strong dose-dependent shift in
subsequent risk estimates (Fig.~\ref{fig:prior_intervention}a). Across
4,574 post-initial snapshots with complete five-condition interventions,
each 10-percentage-point increase in supplied prior risk increased the
updated 24-h ventilation-risk estimate by 3.05 percentage points
(95\% CI, 3.01--3.10), while increasing the supplied prior from 10\% to
90\% shifted the updated prediction by 26.18 percentage points
(95\% CI, 25.74--26.61). Updated predictions were monotonically
non-decreasing across the five prior-risk levels in 70.4\% of snapshots
(95\% CI, 68.9--71.9\%). Thus, despite identical current clinical
evidence, subsequent risk estimates remained systematically sensitive to
the supplied prior belief.

This susceptibility had substantial consequences when the supplied prior
was inconsistent with the observed outcome
(Fig.~\ref{fig:prior_intervention}b; 
Table S6). Outcome-misaligned extreme priors
produced a mean absolute error of 0.588, compared with 0.326 under
outcome-aligned priors, corresponding to an absolute-error increase of 0.262 on the 0--1 probability scale (95\% CI, 0.257--0.266). Because these paired conditions differed only in the supplied prior while holding the current clinical evidence fixed, these results provide direct evidence that supplied prior beliefs causally influence subsequent clinical risk estimates and can increase prediction error when misleading.

The controlled prior-belief effect independently replicated in the hemodynamic deterioration cohort, with current hemodynamic evidence held fixed.
Across 4,440 post-initial snapshots with complete five-condition
interventions, each 10-percentage-point increase in supplied prior risk
increased the updated 24-h vasopressor-risk estimate by 2.89 percentage
points (95\% CI, 2.84--2.93), while increasing the supplied prior from
10\% to 90\% produced a 24.48-percentage-point difference in updated risk
(95\% CI, 24.06--24.89). Updated predictions were monotonically
non-decreasing across the five prior-risk levels in 75.0\% of snapshots
(95\% CI, 73.5--76.5\%). Outcome-misaligned extreme priors increased
absolute prediction error by 0.245 on the 0--1 probability scale
(95\% CI, 0.241--0.249) relative to outcome-aligned priors. The close agreement in effect magnitude across respiratory and hemodynamic endpoints indicates that susceptibility to
supplied prior belief was not specific to mechanical-ventilation
prediction (Fig. S1; Table S6).

\subsection*{Simple prompting does not eliminate harmful longitudinal dependence}
\label{sec:2.12}

Given the causal influence of supplied prior beliefs, we tested two instruction-level mitigation strategies: an evidence-first prompt requiring assessment of current clinical evidence before the previous assessment, and an anti-anchoring prompt warning that the prior assessment could be inaccurate or outdated. Both were evaluated across the 4,580 post-initial snapshots used in the main longitudinal analysis with Qwen3-8B, using independent predictions as the reference.

Neither strategy eliminated the harmful effects associated with
longitudinal context (Table S7).
Evidence-first prompting reduced the mean prior-directed shift by
1.41 percentage points relative to the standard longitudinal prompt
(95\% CI, 1.05--1.77), but did not significantly reduce
harmful-update asymmetry ($\Delta=-0.005$, 95\% CI, $-0.027$ to 0.017) or mean absolute prediction error ($\Delta=0.07$ percentage points, 95\% CI, $-0.23$ to 0.36). Anti-anchoring prompting similarly reduced the mean prior-directed shift
by 1.31 percentage points (95\% CI, 0.95--1.65), but increased the proportion of snapshots harmed relative to independent prediction by 3.95 percentage points
(95\% CI, 2.86--5.08) and increased mean absolute prediction error by
0.39 percentage points (95\% CI, 0.08--0.69).

Thus, explicitly instructing the model to prioritize current evidence or
discount potentially inaccurate prior assessments reduced attraction
toward the prior but did not translate into more reliable longitudinal
updating; anti-anchoring instructions in fact increased both the proportion
of snapshots harmed relative to independent prediction and mean absolute
prediction error.

\subsection*{Evidence validation identifies more reliable longitudinal revisions}
\label{sec:2.13}

We developed a label-free evidence-validated longitudinal update (EVLU) rule to selectively retain longitudinal context when its effect on prediction was supported by changing respiratory evidence. For each post-initial snapshot, we compared the direction of the longitudinal contextual correction---the difference between longitudinal and independent predictions---with respiratory evidence change based on SpO$_2$, FiO$_2$, oxygen flow, respiratory rate, PaO$_2$, PaCO$_2$, and pH, without using outcome labels. The longitudinal prediction was retained only when its contextual correction was non-zero and directionally concordant with respiratory evidence change; otherwise, the prediction reverted to the independent estimate.

Across 4,580 post-initial snapshots, requiring an absolute net respiratory
evidence score of at least one, concordant in direction with the contextual
correction (EVLU-1), reduced the harmful fraction among changed predictions
to 58.42\% and normalized harmful-update asymmetry to 0.168. Requiring an
absolute net respiratory evidence score of at least two with the same
directional concordance criterion (EVLU-2) produced a larger reduction, lowering the harmful fraction to 46.90\% and reversing normalized harmful-update asymmetry to $-0.062$, while mean absolute error differed from independent prediction by only
0.046 percentage points. This gain in update reliability came at substantially
reduced coverage: EVLU-2 retained the longitudinal branch for 5.63\% of
post-initial predictions. Relative to unrestricted longitudinal inference, EVLU-2
reduced the harmful fraction among changed predictions by 13.51 percentage
points (95\% CI, 7.55--19.30) and reduced normalized harmful-update asymmetry
by 0.270 (95\% CI, 0.151--0.386).


To determine whether this greater reliability was explained solely by low
longitudinal-branch retention, we compared EVLU-2 with matched random
retention among the 3,503 nontrivial longitudinal revisions. EVLU-2 retained
258 revisions. Across 10,000 randomizations retaining exactly 258 nontrivial
revisions, the mean harmful fraction was 60.40\%
(95\% randomization interval, 54.65--65.89\%), compared with 46.90\% under
EVLU-2 (empirical $P<0.001$). Revisions retained by EVLU-2 also had a
14.58-percentage-point lower harmful fraction than nontrivial revisions
rejected by the gate (95\% CI, 8.21--20.92) and a 4.23-percentage-point
smaller increase in absolute prediction error (95\% CI, 1.35--7.09).
Thus, the greater reliability of EVLU-2-retained revisions was not explained
by reduced longitudinal-context retention alone.

A stricter absolute net evidence threshold of three (EVLU-3) yielded
numerically more favorable update-quality estimates, with a harmful fraction
of 41.56\% and normalized harmful-update asymmetry of $-0.169$, while
maintaining mean absolute error close to independent prediction. However,
EVLU-3 retained only 1.68\% of post-initial predictions, compared with
5.63\% under EVLU-2, and the differences between EVLU-3 and EVLU-2 were not
statistically distinguishable for harmful fraction, normalized harmful-update
asymmetry, or mean absolute error.
Together, these findings show that longitudinal context need not be
discarded uniformly: restricting its influence to updates supported by
concordant changes in objective clinical evidence improved the reliability
of retained context-induced revisions without requiring outcome labels or
additional model inference, although at substantially reduced coverage.

\begin{figure*}[!htbp]
\centering
\includegraphics[width=\textwidth]{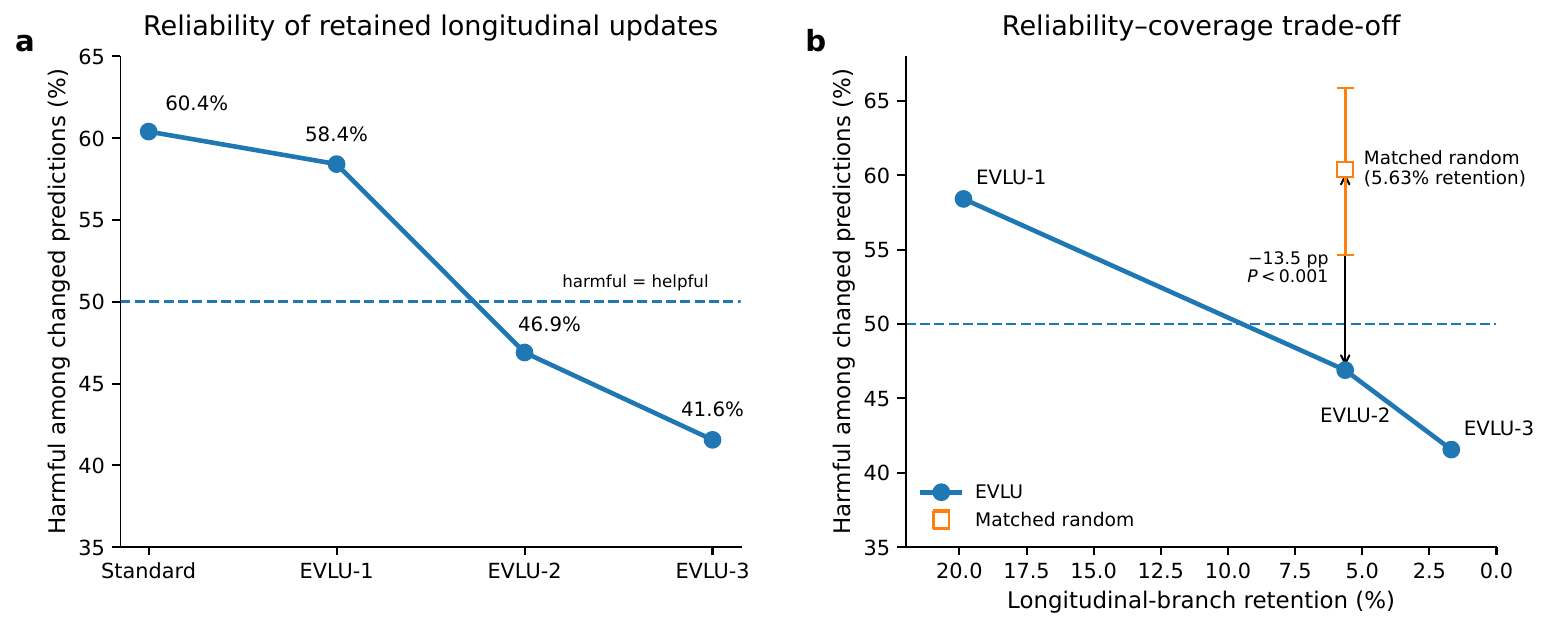}
\caption{
\textbf{Evidence validation identifies a more reliable subset of longitudinal updates.}
\textbf{a}, Harmful fraction among accuracy-changing predictions under
unrestricted longitudinal inference and increasingly stringent EVLU
criteria. The dashed line denotes equal proportions of harmful and helpful
changes (50\%).
\textbf{b}, Reliability--coverage trade-off across EVLU criteria. Stronger evidence requirements reduced the harmful fraction among retained accuracy-changing revisions while progressively reducing longitudinal-branch retention, defined as the proportion of all post-initial predictions for which the longitudinal branch was retained.
}
\label{fig:evlu}
\end{figure*}

\section*{Discussion}
\label{sec:discussion}

In this study, we identify longitudinal belief updating as a distinct reliability challenge for clinical LLMs. Longitudinal self-context more often worsened than improved prediction accuracy when it altered subsequent estimates, a pattern that generalized across model families and replicated for a second clinical endpoint. Controlled interventions revealed asymmetric responsiveness to worsening versus improving evidence and causal susceptibility to supplied prior beliefs despite unchanged current evidence. Prompting reduced prior-directed influence without restoring reliable updating, whereas evidence validation identified a smaller, more reliable subset of revisions. Together, these findings show that evolving patient evidence and model-generated prior assessments should not be treated as interchangeable forms of longitudinal context.

The causal influence of supplied prior assessments is particularly important for longitudinal clinical AI. Previous work has shown that LLM
performance can depend on the ordering and presentation of clinical
information and that models can be sensitive to prior hypotheses and
contextual biases
\cite{hager2024evaluation,mahajan2025cognitive,sheppert2026anchoring}.
Our findings identify a related but distinct problem that arises when a
model's own preceding output becomes part of its subsequent input. In
natural longitudinal trajectories, an association between preceding and
subsequent predictions cannot distinguish the influence of the prior
assessment from that of concurrently evolving patient evidence. The
controlled prior-belief intervention isolates these factors: with current
clinical evidence held fixed, increasing the supplied preceding risk from
10\% to 90\% shifted the subsequent estimate by 26.18 percentage points
for mechanical ventilation and by 24.48 percentage points for vasopressor
initiation. The close agreement across independently constructed respiratory and hemodynamic endpoints indicates that susceptibility to supplied prior belief extends across these two clinically distinct prediction tasks.

This behavior should not be interpreted as a human-like cognitive anchoring mechanism. Rather, it demonstrates causal input--output dependence in which model-generated assessments can influence subsequent risk estimates despite unchanged current evidence. If an inaccurate assessment is propagated as context, it may therefore influence later predictions and allow errors to persist across a longitudinal reasoning chain. 
More broadly, the persistence of harmful longitudinal dependence across model families and scales suggests that increasing model size alone does not consistently attenuate longitudinal unreliability.

A second challenge concerns responses to the direction of changing evidence. Under controlled respiratory interventions, models generally revised risk in the clinically appropriate direction and more strongly as evidence strength increased, indicating recognition of both deterioration and improvement. However, worsening evidence produced larger revisions than matched improving evidence, with deterioration-directed asymmetry persisting at moderate and strong strengths after headroom normalization and across preceding-error and outcome strata.

This distinction matters because reliable longitudinal reasoning requires more than sensitivity to deterioration. Clinical trajectories also involve stabilization, recovery, treatment response, and de-escalation, when elevated risk assessments should be revised downward if supported by new evidence. Dynamic clinical prediction similarly requires repeated updating as time-varying physiological information becomes available \cite{lapp2023dynamic,bull2020repeated,andrinopoulou2021dynamic}. A model that responds more strongly to worsening than comparable improving evidence may therefore be slower to relinquish an elevated assessment as its supporting evidence weakens. Persistence of deterioration-directed asymmetry after headroom normalization and across prediction-error and outcome strata argues against probability-scale constraints, inaccurate preceding predictions, or outcome composition as simple explanations.

The prompting experiments further show that reducing attraction toward a
prior assessment is not equivalent to improving longitudinal belief
updating. Both evidence-first and anti-anchoring instructions reduced the
mean prior-directed shift, but neither improved reliability; anti-anchoring
instead increased both the frequency of harmed snapshots and mean absolute
error. A longitudinal model must determine not simply whether to trust its
previous assessment, but whether the resulting revision is justified by
current evidence. Global instructions to prioritize current evidence or
distrust prior assessments provide no such case-specific validation.
Longitudinal reliability therefore cannot be reduced to a generic
instruction to ``avoid anchoring,'' consistent with evidence that
prompt-based self-correction is often unreliable without informative
external feedback \cite{kamoi2024selfcorrection}.

This distinction motivated our evidence-validation analysis. External or independently derived feedback has been explored to improve model correction \cite{pan2024automatically,kamoi2024selfcorrection}. Evidence-Validated Longitudinal Update (EVLU) tested whether objective changes in patient evidence could identify more reliable context-induced revisions by retaining longitudinal predictions only when their correction direction was concordant with changing respiratory evidence. Increasing the required evidence score produced a reliability--coverage trade-off. With EVLU-2, the harmful fraction among accuracy-changing predictions fell from 60.4\% to 46.9\% and excess mean absolute error relative to independent inference approached zero, while only 5.63\% of post-initial predictions retained the longitudinal branch.

Importantly, the greater reliability of EVLU-2-retained revisions was not explained by low retention alone. Random gates retaining the same number of nontrivial longitudinal revisions preserved a higher harmful fraction and positive harmful-update asymmetry, whereas EVLU-2 shifted the balance toward more helpful than harmful changes. EVLU-2-retained revisions were also more reliable than rejected revisions, indicating that evidence concordance identified more trustworthy longitudinal revisions beyond simply restricting their frequency.

This selectivity nevertheless represents an important limitation rather than a complete solution. A stricter EVLU-3 threshold produced numerically more favorable update-quality measures but retained only 1.68\% of post-initial predictions, without statistically distinguishable improvements over EVLU-2. Even EVLU-2-retained revisions did not improve mean absolute error over independent inference on average. EVLU should therefore be viewed as a conservative proof of concept for evidence-validated belief revision rather than a deployment-ready strategy. Future methods will need to preserve this selectivity while substantially increasing useful longitudinal coverage.

These findings have broader implications for clinical AI systems that operate repeatedly over time. Snapshot-level evaluation cannot reveal whether earlier model-generated assessments improve, distort, or constrain subsequent reasoning. Although longitudinal clinical AI increasingly represents extended patient histories and temporal relationships \cite{kruse2025large,li2026clicare}, access to longitudinal information does not ensure reliable longitudinal reasoning. Model-generated history may provide continuity but must remain revisable in light of new patient evidence.

Evidence validation provides one possible design principle: longitudinal systems could test whether a context-induced revision is supported by independently evolving clinical evidence before propagating it forward. This selective use of context parallels approaches that trade coverage for reliability by deferring insufficiently trustworthy predictions \cite{dvijotham2023deferral}. Longitudinal evaluations should therefore assess not only discrimination and calibration, but also the direction, magnitude, evidence dependence, and consequences of belief revision over time \cite{hager2024evaluation,rao2026large,kruse2025large}.

Several limitations should be considered. First, both endpoint-specific cohorts were derived from MIMIC-IV. Although the natural longitudinal failure pattern and controlled prior-belief susceptibility replicated across respiratory and hemodynamic deterioration cohorts, generalizability to other institutions, populations, clinical tasks, and timescales remains to be established. Second, trajectories were constructed primarily from structured clinical variables; notes, imaging, treatment decisions, and other real-world information could alter evidence integration and dependence on preceding assessments. Third, the controlled respiratory perturbations were designed to isolate directional evidence updating and should not be interpreted as naturally occurring counterfactual trajectories. Fourth, although harmful longitudinal updating was consistent across models, the mechanistic, prompting, and evidence-validation experiments focused on the primary model and require replication across model families.
Fifth, absolute discrimination was modest. Our analysis was designed to isolate within-snapshot belief revision under matched inference conditions rather than optimize clinical risk prediction; whether these updating failures persist in clinically optimized or task-adapted models remains to be established.
Sixth, EVLU relies on manually specified respiratory evidence rules and provides a conservative rather than comprehensive solution. Although matched-retention analyses showed that its reliability gains were not attributable solely to reduced longitudinal-context use, more reliable operating points retained only a small fraction of longitudinal revisions. EVLU therefore establishes evidence validation as an interpretable design principle rather than a clinically validated gating system. Future work should evaluate evidence validation across independent datasets and endpoints, incorporate multimodal and unstructured evidence, and develop learned or task-adaptive strategies that increase useful coverage without reintroducing unsupported prior dependence. Prospective and clinician-centered evaluation will ultimately be necessary before clinical deployment, consistent with evidence that real-world clinical utility cannot be inferred from model-level performance alone \cite{leibovitch2026prospective,agweyu2026generative}.

In summary, longitudinal belief updating represents a distinct reliability challenge for clinical LLMs. Supplied prior assessments can causally influence subsequent risk estimates despite unchanged current evidence, while evolving evidence produces asymmetric revision between worsening and improving states. Prompting did not restore longitudinal reliability. Evidence validation preferentially identified more reliable revisions beyond reduced longitudinal-context retention alone, although at substantial cost to coverage. Clinical LLMs intended for repeated use should therefore be evaluated not only on what they predict, but also on how, when, and on what evidence they revise preceding assessments.

\section*{Methods}\label{sec:methods}
\subsection*{Data source and study cohort}

We used MIMIC-IV version 3.1~\cite{johnson2023mimic}, a publicly available de-identified electronic health record dataset containing hospital and intensive care unit (ICU) data. We considered adult ICU stays from patients aged 18 years or older and required a minimum ICU length of stay of 30 hours to ensure sufficient longitudinal observation. The study focused on evolving clinical deterioration during the early ICU course, with invasive mechanical ventilation as the primary endpoint and continuous vasopressor initiation as an independently constructed
hemodynamic replication endpoint. For the vasopressor replication analysis, the corresponding structured fields described hemodynamic assessment and 24-hour vasopressor risk,
while all other aspects of the independent and longitudinal inference
protocol were unchanged. The creation and sharing of MIMIC-IV were approved by the Institutional
Review Board of Beth Israel Deaconess Medical Center, with a waiver of
informed consent. The present study used only de-identified data available
through PhysioNet.

We constructed a matched cohort consisting of ICU stays that subsequently underwent invasive mechanical ventilation and control stays without qualifying ventilation events. Invasive ventilation events were identified from structured ICU records
using prespecified rules for invasive airway devices, ventilation modes,
oxygen-delivery fields, and procedure labels, while explicitly excluding
non-invasive respiratory-support modalities. For each ICU stay, the first
qualifying invasive ventilation event was defined as the ventilation onset
time. 

\subsection*{Invasive ventilation outcome and matched cohort construction}

Ventilation-positive ICU stays were required to have their first invasive ventilation event between 12 and 72 hours after ICU admission. This interval ensured that sufficient clinical history was available before ventilation while restricting the analysis to early ICU deterioration. Candidate positive stays were further required to contain at least two scheduled clinical snapshots occurring before ventilation onset.

Clinical snapshots were scheduled at 6, 12, 18, and 24 hours after ICU admission. For ventilation-positive stays, only snapshots occurring strictly before the first invasive ventilation event were retained. At each snapshot, the prediction target was whether invasive mechanical ventilation would be initiated within the subsequent 24 hours.

We matched ventilation-positive stays 1:1 to negative ICU stays without
replacement. Matching required the same sex and first ICU care unit, an age
difference of no more than 10 years, and sufficient ICU duration for the
negative stay to support the complete snapshot schedule of its matched
positive case together with the subsequent 24-hour prediction horizon.
Among 65,366 ICU stays screened, 2,025 met the initial positive eligibility
criteria and 24,998 were eligible negative controls. After requiring at least
two pre-ventilation snapshots, 2,023 positive stays remained. We selected
1,000 matched pairs, yielding 2,000 unique ICU stays corresponding to 2,000
unique patients. Seven positive candidates considered during matching could
not be assigned an eligible control before the target cohort size was reached.
Among the final matched pairs, the mean absolute age difference was 0.02 years
and the maximum difference was 4 years.

\subsection*{Longitudinal clinical trajectory construction}

For each matched ICU stay, we constructed a longitudinal trajectory from structured EHR measurements available up to each scheduled snapshot. The clinical state included vital signs and respiratory measurements, including oxygen saturation, respiratory rate, heart rate, systolic and diastolic blood pressure, temperature, fraction of inspired oxygen, and supplemental oxygen flow, together with laboratory measurements including arterial oxygen and carbon dioxide pressure, pH, lactate, white blood cell count, bicarbonate, and creatinine.

For each snapshot at time $t$, we represented two complementary views of the patient's clinical state. The \emph{current state} summarized measurements available within the preceding 24 hours, whereas \emph{new evidence} contained measurements and respiratory-support events observed since the immediately preceding snapshot. Numerical measurements were summarized using the most recent value together with the observed minimum, maximum, median, and measurement count within the corresponding interval.

The binary snapshot-level outcome was defined as

\[
y_t =
\mathbb{I}
\left(
t < T_{\mathrm{vent}}
\leq t + 24\mathrm{h}
\right),
\]

where $T_{\mathrm{vent}}$ denotes the first invasive ventilation time. All snapshots from ventilation-positive patients were verified to occur strictly before ventilation onset. The resulting cohort contained 6,580 longitudinal snapshots, with 3,290 snapshots from ventilation-positive patients and 3,290 from their matched controls.

\subsection*{Independent vasopressor cohort}

To evaluate generalization across clinical deterioration domains, we
independently constructed a second endpoint-specific cohort for continuous
vasopressor initiation. Vasopressor exposure was identified from MIMIC-IV
\texttt{inputevents} using continuous medication records for dopamine,
epinephrine, norepinephrine, phenylephrine, and vasopressin. Drug-push
administrations were excluded. For each ICU stay, the first qualifying
continuous vasopressor administration was defined as the vasopressor
initiation time, and the prediction target at each snapshot was whether
continuous vasopressor support would be initiated within the subsequent
24 hours.

Cohort construction paralleled the invasive mechanical ventilation
analysis. Vasopressor-positive stays were required to have their first
continuous vasopressor initiation between 12 and 72 hours after ICU
admission and at least two scheduled snapshots strictly preceding
vasopressor initiation. Negative stays had no detected continuous
vasopressor administration during the ICU stay. Among 44,573 adult ICU
stays satisfying the length-of-stay criterion, 2,352 met the positive
endpoint criteria and 28,049 were eligible strict negative controls.
Positive stays were independently matched 1:1 without replacement to
negative stays, requiring the same sex and first ICU care unit, an age
difference of no more than 10 years, and sufficient follow-up in the
negative stay to cover the corresponding snapshot schedule and subsequent
24-hour prediction horizon. We randomly selected 1,000 eligible vasopressor-positive stays and matched each 1:1 to an eligible negative stay without replacement, yielding 1,000 matched pairs and 2,000 unique ICU stays corresponding to 2,000 unique patients.

Clinical snapshots were scheduled at 6, 12, 18, and 24 hours after ICU
admission, as in the primary analysis. For the vasopressor endpoint,
structured clinical evidence emphasized hemodynamic state and included
heart rate, systolic and diastolic blood pressure, mean arterial pressure,
lactate, urine output, creatinine, bicarbonate, and white blood cell count.
The independent and longitudinal evaluation conditions otherwise followed
the same protocol as the invasive mechanical ventilation analysis, with
the longitudinal condition additionally receiving the model's structured
assessment from the preceding snapshot.

\subsection*{Independent and longitudinal LLM evaluation}

We evaluated clinical belief updating under two matched inference conditions. In the \emph{independent} condition, the model assessed each clinical snapshot independently using only the patient information available up to the current time point. The model therefore had no access to its own assessment from the preceding snapshot.

In the \emph{longitudinal} condition, the model received the same current clinical information together with its structured assessment from the immediately preceding snapshot. The prior assessment included the previous respiratory assessment, predicted probability of invasive ventilation within 24 hours, and evidence identified for and against respiratory deterioration. At each subsequent snapshot, the model was instructed to either maintain or revise its previous assessment according to the newly available clinical evidence.

For the mechanical-ventilation endpoint, the model produced a structured
response containing a respiratory assessment, a predicted probability of
invasive ventilation within the subsequent 24 hours, evidence supporting
and arguing against deterioration, and a \texttt{KEEP} or \texttt{REVISE}
decision when prior context was available. The vasopressor replication used
the corresponding endpoint-specific schema with a hemodynamic assessment,
a predicted probability of continuous vasopressor initiation within the
subsequent 24 hours, supporting and opposing hemodynamic evidence, and the
same \texttt{KEEP} or \texttt{REVISE} decision structure. We used Qwen3-8B as the primary model and performed deterministic local inference with sampling disabled and visible chain-of-thought generation disabled.

For each snapshot after the first observation, we quantified the effect of prior model context as

\[
\Delta_t^{\mathrm{context}}
=
P_t^{\mathrm{long}}
-
P_t^{\mathrm{ind}},
\]

where $P_t^{\mathrm{long}}$ and $P_t^{\mathrm{ind}}$ denote the longitudinal and independent ventilation-risk estimates, respectively. We additionally compared absolute prediction error, Brier score, and discrimination between the two conditions.

\subsection*{Cross-model evaluation}

To assess whether harmful longitudinal dependence generalized across model
families and scales, we repeated the independent and longitudinal evaluation
on the same 2,000-patient mechanical-ventilation cohort using Qwen3-32B,
Llama-3.1-8B-Instruct, Mistral-7B-Instruct-v0.3,
Mistral-Small-3.1-24B-Instruct-2503, MedGemma-4B-IT,
MedGemma-27B-Text-IT, and Gemma-3-12B-IT, in addition to the primary
Qwen3-8B model. All models received the same clinical snapshots, prediction
target, and independent-versus-longitudinal prompt structure. All models
were evaluated locally with deterministic decoding using their model-specific
chat templates.

\subsection*{Controlled respiratory evidence interventions}

To distinguish systematic evidence-responsive belief updating from changes arising from naturally evolving patient trajectories, we constructed a controlled counterfactual experiment. For each eligible transition between consecutive longitudinal snapshots, we fixed both the patient's preceding objective respiratory state and the model's preceding longitudinal assessment, and varied only the newly presented respiratory evidence.

We generated six directional interventions defined by two evidence directions (\emph{worsening} and \emph{improving}) and three evidence strengths (\emph{subtle}, \emph{moderate}, and \emph{strong}). Interventions were patient specific and were generated relative to the observed respiratory state at the preceding snapshot. For worsening interventions, respiratory rate and supplemental oxygen requirements were increased while oxygen saturation was decreased; improving interventions applied the corresponding changes in the opposite direction. The subtle, moderate, and strong interventions altered respiratory rate by 4, 8, and 14 breaths/min, oxygen saturation by 2, 4, and 7 percentage points, and supplemental oxygen flow by 1, 2, and 4 L/min, respectively,
subject to predefined physiological bounds. Intervention direction and magnitude were defined independently of the observed clinical outcome.

Two control conditions were additionally constructed. In the \emph{redundant} condition, the newly presented evidence explicitly restated the preceding respiratory measurements without changing their values. In the \emph{neutral} condition, the model was informed that no new respiratory measurements or directional evidence were available. These controls were designed to determine whether belief revision reflected clinically meaningful evidence rather than generic sensitivity to the presence of additional prompt content.

Transitions were included in the controlled experiment when both respiratory rate and oxygen saturation were available at the preceding snapshot. Of 4,580 real longitudinal transitions, 4,545 (99.2\%) satisfied these criteria. Each transition was evaluated under all eight counterfactual conditions, producing 36,360 model responses in a fully paired design.

\subsection*{Controlled prior-belief interventions}
\label{sec:methods_prior_intervention}

To isolate the causal influence of prior model beliefs from changes in
current patient evidence, we performed controlled prior-belief interventions
on all post-initial snapshots. For each snapshot, the current clinical
evidence and corresponding user prompt were held identical across
intervention conditions, while only the numerical risk estimate supplied
in the previous model assessment was manipulated. We evaluated five
synthetic prior-risk conditions: 10\%, 30\%, 50\%, 70\%, and 90\%. For
the mechanical-ventilation endpoint, the remaining fields of the previous
assessment consisted of an empty respiratory assessment and empty lists of
evidence for and against deterioration. For the vasopressor endpoint, the
corresponding hemodynamic assessment and evidence fields were likewise held
empty and identical across conditions. Thus, within each endpoint, the
intervention varied only the numerical prior belief while preserving the
current patient state and all other prompt content.

For the mechanical-ventilation cohort, each of the 4,580 post-initial
snapshots was evaluated under all five prior-risk conditions using the same
model and inference configuration as in the primary longitudinal experiment,
yielding 22,900 model responses. Of these, 4,574 snapshots had valid
predictions under all five intervention conditions and were included in
complete-case dose--response analyses. We repeated the identical intervention
protocol in the independently constructed vasopressor cohort. Each of 4,444
post-initial snapshots was evaluated under the same five prior-risk
conditions, yielding 22,220 model responses; 4,440 snapshots had valid
predictions under all five conditions and were retained for complete-case
analyses.

For each endpoint, we quantified prior susceptibility as the mean change in
the subsequent risk prediction associated with a 10-percentage-point increase
in supplied prior risk. We additionally measured the within-snapshot
difference between predictions under the 90\% and 10\% prior conditions and
the proportion of snapshots for which updated risk was monotonically
non-decreasing across the five prior-risk levels.

To evaluate the consequences of correct versus misleading prior beliefs, we
classified the extreme prior conditions according to their alignment with
the observed binary outcome. For outcome-positive snapshots, the 90\% prior
was treated as outcome-aligned and the 10\% prior as outcome-misaligned;
for outcome-negative snapshots, the 10\% prior was treated as outcome-aligned
and the 90\% prior as outcome-misaligned. Absolute prediction error was
compared between these paired conditions. Because current clinical evidence
was held fixed within each snapshot and only the supplied prior risk was
experimentally manipulated, differences in subsequent predictions across
intervention conditions isolate the effect of the supplied prior belief.

\subsection*{Prompt-based mitigation}
\label{sec:methods_prompt_mitigation}

To test whether instruction-level interventions could reduce harmful
dependence on previous model assessments, we evaluated two alternative
longitudinal prompting strategies using Qwen3-8B. The
\emph{evidence-first} strategy instructed the model to first assess the
current clinical snapshot independently and form an initial
ventilation-risk estimate before reviewing the preceding model assessment
as historical context. The \emph{anti-anchoring} strategy explicitly
instructed the model that the preceding assessment could be inaccurate,
outdated, or overly confident and should not be treated as evidence about
the current patient state. Both prompts emphasized that the current
clinical evidence should take precedence when it conflicted with the
preceding assessment.

Each strategy was evaluated on all 4,580 post-initial snapshots using the
same current clinical information, preceding model assessments, output
format, and deterministic inference configuration as the standard
longitudinal condition. The independent current-snapshot predictions were
unchanged and served as the common reference condition. We compared each
prompting strategy with the standard longitudinal prompt using the
proportion of snapshots harmed by longitudinal context, the harmful
fraction among accuracy-changing predictions, normalized harmful-update
asymmetry, change in mean absolute prediction error relative to independent
inference, and the magnitude of prior-directed movement.

\subsection*{Evidence-validated longitudinal updating}
\label{sec:methods_evlu}

To test whether harmful longitudinal dependence could be mitigated by
requiring support from evolving patient evidence, we developed an
Evidence-Validated Longitudinal Update (EVLU) strategy. EVLU is a
rule-based post hoc validation procedure and requires neither additional
model training nor additional LLM inference. For each post-initial
snapshot $t$, we defined the context-induced longitudinal correction as
\begin{equation}
\Delta_t^{\mathrm{ctx}}
=
P_t^{\mathrm{long}}-P_t^{\mathrm{ind}},
\end{equation}
where $P_t^{\mathrm{ind}}$ is the ventilation-risk estimate obtained from
the current snapshot alone and $P_t^{\mathrm{long}}$ is the estimate
obtained when the previous model assessment is additionally provided.

We independently characterized the direction of change in structured
respiratory evidence between consecutive snapshots. Seven respiratory
variables were considered when available: SpO$_2$, FiO$_2$, oxygen flow,
respiratory rate, PaO$_2$, PaCO$_2$, and pH. Each variable contributed a
directional vote $v_{tj}\in\{-1,0,+1\}$, representing improvement, no
meaningful change, or deterioration, respectively. A vote was assigned
only when the change exceeded a prespecified minimum threshold: 2
percentage points for SpO$_2$, 5 percentage points for FiO$_2$, 1 L/min
for oxygen flow, 3 breaths/min for respiratory-rate abnormality, 10 mmHg
for PaO$_2$, 5 mmHg for PaCO$_2$, and 0.03 for pH. Decreases in
SpO$_2$, PaO$_2$, and pH and increases in FiO$_2$, oxygen flow, and
PaCO$_2$ were treated as evidence of deterioration; respiratory rate was
scored according to increasing deviation from the reference range of
12--20 breaths/min. Changes in the opposite direction were treated as
evidence of improvement. Variables unavailable at either snapshot did
not contribute a directional vote.

The net evidence score was
\begin{equation}
C_t=\sum_j v_{tj},
\end{equation}
with positive values indicating net evidence of respiratory
deterioration and negative values indicating net evidence of improvement.
For a concordance threshold $k$, a non-zero longitudinal correction was
retained only when the absolute net evidence score was at least $k$ and
its direction agreed with the direction of the context-induced correction:
\begin{equation}
w_t^{(k)}
=
\mathbb{I}
\left(
|C_t|\geq k
\;\land\;
\operatorname{sign}(C_t)
=
\operatorname{sign}(\Delta_t^{\mathrm{ctx}})
\right).
\end{equation}
The resulting EVLU prediction was
\begin{equation}
P_t^{\mathrm{EVLU}-k}
=
P_t^{\mathrm{ind}}
+
w_t^{(k)}
\left(
P_t^{\mathrm{long}}-P_t^{\mathrm{ind}}
\right).
\end{equation}
Thus, when a non-zero longitudinal revision was supported by a sufficiently
large net respiratory evidence score in the same direction, EVLU retained
the longitudinal prediction; otherwise, it reverted to the
current-snapshot-only prediction.

We evaluated concordance thresholds $k=1,2,3$ to characterize the
trade-off between retaining longitudinal information and selectively
accepting more strongly corroborated revisions. These thresholds were
evaluated as a sensitivity analysis rather than selected by optimizing
outcome performance. No outcome labels were used to construct the
evidence score or determine whether an individual longitudinal revision
was retained. We compared EVLU with unrestricted longitudinal inference and
independent current-snapshot inference.
Primary mitigation outcomes were the fraction of retained changed
predictions that increased absolute prediction error, normalized
harmful-update asymmetry, change in mean absolute error relative to
independent inference, and longitudinal-update retention, defined as the
proportion of all post-initial predictions for which the longitudinal branch
was retained. Confidence intervals for paired policy comparisons were estimated using
stay-cluster bootstrap resampling with 5,000 replicates.

\subsection*{Evaluation metrics}

We evaluated model behavior from both predictive-performance and belief-updating perspectives.

For the natural longitudinal evaluation, predictive performance was assessed using the area under the receiver operating characteristic curve (AUROC), Brier score, and mean absolute error (MAE) between the predicted probability of invasive ventilation and the binary 24-hour outcome. To quantify the effect of prior model context, we computed the paired prediction difference at each snapshot after the first observation as

\[
\Delta_t^{\mathrm{context}}
=
P_t^{\mathrm{long}}
-
P_t^{\mathrm{ind}},
\]

where $P_t^{\mathrm{long}}$ and $P_t^{\mathrm{ind}}$ denote the longitudinal and independent ventilation-risk estimates, respectively. We additionally report the absolute condition difference,

\[
\left|
P_t^{\mathrm{long}}
-
P_t^{\mathrm{ind}}
\right|,
\]

to quantify the magnitude of belief change induced by access to the prior assessment.

To determine whether prior context improved or worsened prediction accuracy, we calculated the change in absolute prediction error,

\[
\Delta E_t
=
\left|
P_t^{\mathrm{long}}-y_t
\right|
-
\left|
P_t^{\mathrm{ind}}-y_t
\right|,
\]

where $y_t\in\{0,1\}$ is the snapshot-level ventilation outcome. A negative
value indicates that longitudinal context improved prediction accuracy,
whereas a positive value indicates that it increased prediction error. We
accordingly categorized snapshots as \emph{helped}, \emph{hurt}, or
\emph{no effect}. Among snapshots with a non-zero change in absolute
prediction error, the harmful fraction was defined as
\[
F_{\mathrm{harm}}
=
\frac{N_{\mathrm{hurt}}}
{N_{\mathrm{hurt}}+N_{\mathrm{helped}}},
\]
and normalized harmful-update asymmetry was defined as
\[
A_{\mathrm{harm}}
=
\frac{N_{\mathrm{hurt}}-N_{\mathrm{helped}}}
{N_{\mathrm{hurt}}+N_{\mathrm{helped}}}.
\]
Thus, $A_{\mathrm{harm}}>0$ indicates that harmful effects were more frequent
than helpful effects among accuracy-changing predictions.

We quantified prior-belief attraction using a prior-directed shift measure, which was also used in the prompting analyses. This measure was defined as
\[
\mathrm{PriorPull}_t
=
\left|
P_t^{\mathrm{ind}}
-
P_{t-1}^{\mathrm{long}}
\right|
-
\left|
P_t^{\mathrm{long}}
-
P_{t-1}^{\mathrm{long}}
\right|.
\]

Positive values indicate that the longitudinal prediction remained closer to the model's preceding belief than the independent prediction based on the current clinical state alone.

For the controlled counterfactual experiment, belief revision was quantified as

\[
\Delta P
=
P_{\mathrm{updated}}
-
P_{\mathrm{prior}},
\]

where $P_{\mathrm{prior}}$ is the model's risk estimate before the controlled evidence intervention and $P_{\mathrm{updated}}$ is the corresponding estimate after incorporating the intervention. For worsening evidence, positive $\Delta P$ represents directionally appropriate updating; for improving evidence, negative $\Delta P$ represents directionally appropriate updating. We report the mean and median change in risk, mean absolute change, and the proportions of positive, negative, and unchanged revisions.

To characterize evidence-strength sensitivity, we evaluated dose-response behavior across subtle, moderate, and strong interventions. A transition was considered monotonic when the magnitude of belief revision was non-decreasing across increasing evidence strengths.

Directional asymmetry was quantified on the raw probability scale as

\[
A_{\mathrm{raw}}
=
\Delta P_{\mathrm{worsening}}
+
\Delta P_{\mathrm{improving}},
\]

because $\Delta P_{\mathrm{improving}}$ is expected to be negative. Thus, $A_{\mathrm{raw}}=0$ represents equal-magnitude updating in opposite directions, whereas $A_{\mathrm{raw}}>0$ indicates stronger deterioration-directed updating.

Because probability estimates are bounded between 0 and 100, the available upward and downward revision ranges depend on the prior risk. We therefore additionally calculated headroom-normalized directional responses:

\[
R_{\mathrm{worsening}}
=
\frac{
\Delta P_{\mathrm{worsening}}
}{
100-P_{\mathrm{prior}}
},
\]

and

\[
R_{\mathrm{improving}}
=
\frac{
-\Delta P_{\mathrm{improving}}
}{
P_{\mathrm{prior}}
}.
\]

The normalized asymmetry score was defined as

\[
A_{\mathrm{norm}}
=
R_{\mathrm{worsening}}
-
R_{\mathrm{improving}}.
\]

Positive $A_{\mathrm{norm}}$ indicates stronger deterioration-directed belief updating after accounting for the unequal upward and downward probability headroom.

\subsection*{Statistical analysis}

All primary comparisons were paired at the clinical-snapshot or transition
level, as appropriate. Because multiple observations could be contributed by
the same ICU stay, uncertainty was quantified using cluster bootstrap
resampling at the ICU-stay level to preserve within-stay dependence. Unless
otherwise specified, 95\% confidence intervals were estimated using 10,000
ICU-stay-level bootstrap replicates.

For the natural longitudinal evaluation, independent and longitudinal
predictions were compared at matched clinical snapshots. Predictive
performance was summarized using AUROC, Brier score, and mean absolute error
(MAE). For post-initial snapshots, we additionally summarized the paired
difference between longitudinal and independent risk estimates and classified
context-induced changes as harmful, helpful, or having no effect according to
their effect on absolute prediction error. Cross-model uncertainty for
harmful-update asymmetry was estimated by ICU-stay-level bootstrap resampling.

For the controlled respiratory evidence interventions, all worsening and
improving conditions were paired within the same longitudinal transition.
Directional asymmetry was evaluated on both the raw probability scale and
after normalization by the available upward or downward probability headroom,
as defined above. We additionally examined headroom-normalized asymmetry
across strata of preceding predicted risk, preceding prediction error, and
clinical outcome. 

For the controlled prior-belief interventions, all prior-risk conditions were
paired within the same clinical snapshot while current clinical evidence was
held fixed. Prior susceptibility was quantified from the within-snapshot
dose--response relationship across the five supplied prior-risk levels.
Uncertainty for the mean effect of a 10-percentage-point increase in supplied
prior risk, the 90\%-versus-10\% contrast, monotonicity, and the comparison
between outcome-aligned and outcome-misaligned priors was estimated by
ICU-stay-level bootstrap resampling.

For the prompt-based mitigation analyses, each alternative longitudinal prompt
was compared with the standard longitudinal prompt using the same post-initial
snapshots and the same independent predictions as the common reference.
Differences in harmful-update measures, mean absolute error, and
prior-directed shift were evaluated using paired ICU-stay-level bootstrap
resampling with 5,000 replicates.

For EVLU, policy-level outcomes were compared with unrestricted longitudinal
inference and independent current-snapshot inference using paired
ICU-stay-level bootstrap resampling with 5,000 replicates. To determine whether the greater reliability of EVLU-2-retained revisions could be explained solely by low longitudinal-branch retention, we performed a matched-random retention analysis among the 3,503 nontrivial longitudinal revisions. EVLU-2 retained 258 such revisions. We therefore generated 10,000 random assignments, each retaining exactly 258 of the 3,503 nontrivial revisions, and recomputed the
harmful fraction, normalized harmful-update asymmetry, and change in mean
absolute error for each assignment.
The resulting randomization distributions were summarized using the
2.5th and 97.5th percentiles as 95\% randomization intervals. The one-sided
empirical randomization $P$ value was calculated as
$(1+b)/(B+1)$, where $B=10{,}000$ and $b$ was the number of random
assignments yielding a harmful fraction at least as low as that observed
under EVLU-2. We additionally compared EVLU-2-retained and rejected
nontrivial revisions using ICU-stay-level cluster bootstrap resampling,
with 10,000 replicates, for harmful fraction, normalized harmful-update
asymmetry, and mean change in absolute prediction error.

Primary analyses emphasized paired effect estimates with confidence
intervals. No adjustment for multiple comparisons was applied to exploratory
subgroup or sensitivity analyses, which were interpreted descriptively
rather than as independent confirmatory hypothesis tests. Analyses were
performed using Python.

\section*{Data Availability}
The clinical data used in this study were obtained from MIMIC-IV, a publicly
available de-identified electronic health record database. Access to MIMIC-IV
is available through PhysioNet to credentialed users who complete the required
training and data-use agreement. The derived cohorts used in this study can be
reconstructed from MIMIC-IV using the cohort-construction procedures described
in the Methods and accompanying code.

\section*{Code Availability}

Code supporting this study, including scripts for cohort construction,
longitudinal model evaluation, controlled interventions, and statistical
analysis, is available to editors and reviewers through a private Zenodo reviewer-access link. The repository contains the scripts required to reproduce the analyses reported in this study from credentialed access to MIMIC-IV and will be made publicly available upon publication.

\section*{Acknowledgements}
None.

\section*{Funding}
This work was supported by the National Center for Complementary and Integrative Health, United States [grant numbers R01AT009457, U01AT012871]; the National Institute on Aging, United States [grant number R01AG078154]; the National Cancer Institute, United States [grant number R01CA287413]; the National Institute of Diabetes and Digestive and Kidney Diseases, United States [grant number R01DK115629]; the National Institute on Minority Health and Health Disparities, United States [grant number R21MD019134]; and the Food and Drug Administration, United States [grant number U01FD008720]. The funders had no role in the study design, data collection and analysis, decision to publish, or preparation of the manuscript.

\section*{Author Contributions}
M.Z. conceived the study, developed the methodology, implemented the
computational pipeline, performed the experiments and statistical analyses,
interpreted the results, prepared the figures, and drafted the manuscript.
R.Z. contributed to study conception and design, supervised the work,
provided resources and funding, contributed to interpretation of the results,
and reviewed and edited the manuscript. Both authors reviewed and approved
the final manuscript.

\section*{Competing Interests}
The authors declare no competing interests.


\bibliography{sn-bibliography}

@article{johnson2023mimic,
  title={MIMIC-IV, a freely accessible electronic health record dataset},
  author={Johnson, Alistair EW and Bulgarelli, Lucas and Shen, Lu and Gayles, Alvin and Shammout, Ayad and Horng, Steven and Pollard, Tom J and Hao, Sicheng and Moody, Benjamin and Gow, Brian and others},
  journal={Scientific Data},
  volume={10},
  number={1},
  pages={1},
  year={2023},
  publisher={Nature Publishing Group},
  doi={10.1038/s41597-022-01899-x}
}

@article{singhal2023clinical,
  title   = {Large language models encode clinical knowledge},
  author  = {Singhal, Karan and Azizi, Shekoofeh and Tu, Tao and Mahdavi, S. Sara and Wei, Jason and Chung, Hyung Won and Scales, Nathan and Tanwani, Ajay and Cole-Lewis, Heather and Pfohl, Stephen and Payne, Perry and Seneviratne, Martin and Gamble, Paul and Kelly, Chris and Babiker, Abubakr and Sch{\"a}rli, Nathanael and Chowdhery, Aakanksha and Mansfield, Philip and Demner-Fushman, Dina and Ag{\"u}era y Arcas, Blaise and Webster, Dale and Corrado, Greg S. and Matias, Yossi and Chou, Katherine and Gottweis, Juraj and Tomasev, Nenad and Liu, Yun and Rajkomar, Alvin and Barral, Joelle and Semturs, Christopher and Karthikesalingam, Alan and Natarajan, Vivek},
  journal = {Nature},
  volume  = {620},
  number  = {7972},
  pages   = {172--180},
  year    = {2023},
  doi     = {10.1038/s41586-023-06291-2}
}

@article{bedi2025testing,
  title   = {Testing and Evaluation of Health Care Applications of Large Language Models: A Systematic Review},
  author  = {Bedi, Suhana and Liu, Yutong and Orr-Ewing, Lucy and Dash, Dev and Koyejo, Sanmi and Callahan, Alison and Fries, Jason A. and Wornow, Michael and Swaminathan, Akshay and Lehmann, Lisa Soleymani and Hong, Hyo Jung and Kashyap, Mehr and Chaurasia, Akash R. and Shah, Nirav R. and Singh, Karandeep and Tazbaz, Troy and Milstein, Arnold and Pfeffer, Michael A. and Shah, Nigam H.},
  journal = {JAMA},
  volume  = {333},
  number  = {4},
  pages   = {319--328},
  year    = {2025},
  doi     = {10.1001/jama.2024.21700}
}

@article{hager2024evaluation,
  title   = {Evaluation and mitigation of the limitations of large language models in clinical decision-making},
  author  = {Hager, Paul and Jungmann, Friederike and Holland, Robbie and Bhagat, Kunal and Hubrecht, Inga and Knauer, Manuel and Vielhauer, Jakob and Makowski, Marcus and Braren, Rickmer and Kaissis, Georgios and Rueckert, Daniel},
  journal = {Nature Medicine},
  volume  = {30},
  number  = {9},
  pages   = {2613--2622},
  year    = {2024},
  doi     = {10.1038/s41591-024-03097-1}
}

@article{wilie2024belief,
  title   = {Belief Revision: The Adaptability of Large Language Models Reasoning},
  author  = {Wilie, Bryan and Cahyawijaya, Samuel and Ishii, Etsuko and He, Junxian and Fung, Pascale},
  journal = {Proceedings of the 2024 Conference on Empirical Methods in Natural Language Processing},
  pages   = {10480--10496},
  year    = {2024},
  doi     = {10.18653/v1/2024.emnlp-main.586}
}

@article{mahajan2025cognitive,
  title   = {Cognitive bias in clinical large language models},
  author  = {Mahajan, Arjun and Obermeyer, Ziad and Daneshjou, Roxana and Lester, Jenna and Powell, Dylan},
  journal = {npj Digital Medicine},
  volume  = {8},
  pages   = {428},
  year    = {2025},
  doi     = {10.1038/s41746-025-01790-0}
}

@article{singhal2025expert,
  title   = {Toward expert-level medical question answering with large language models},
  author  = {Singhal, Karan and Tu, Tao and Gottweis, Juraj and Sayres, Rory and Wulczyn, Ellery and Amin, Mohamed and Hou, Le and Clark, Kevin and Pfohl, Stephen R. and Cole-Lewis, Heather and Neal, Darlene and Rashid, Qazi Mamunur and Schaekermann, Mike and Wang, Amy and Dash, Dev and Chen, Jonathan H. and Shah, Nigam H. and Lachgar, Sami and Mansfield, Philip Andrew and Prakash, Sushant and Green, Bradley and Dominowska, Ewa and Ag{\"u}era y Arcas, Blaise and Toma{\v{s}}ev, Nenad and Liu, Yun and Wong, Renee and Semturs, Christopher and Mahdavi, S. Sara and Barral, Joelle K. and Webster, Dale R. and Corrado, Greg S. and Matias, Yossi and Azizi, Shekoofeh and Karthikesalingam, Alan and Natarajan, Vivek},
  journal = {Nature Medicine},
  volume  = {31},
  number  = {3},
  pages   = {943--950},
  year    = {2025},
  doi     = {10.1038/s41591-024-03423-7}
}

@article{rao2026large,
  title   = {Large Language Model Performance and Clinical Reasoning Tasks},
  author  = {Rao, Arya S. and Esmail, Kaiz P. and Lee, Richard S. and Jiang, Sharon and Arraiza Carlo, Bianca and Gill, Jasleen and Khanna, Praneet and Kalmowitz, Ezra and Montagnese, Basile and Heydari, Kimia and Jiao, Qiao and Bott, Ethan and Nguyen, Dan and Wang, Grace and Hood, Michael and Landman, Adam B. and Succi, Marc D.},
  journal = {JAMA Network Open},
  volume  = {9},
  number  = {4},
  pages   = {e264003},
  year    = {2026},
  doi     = {10.1001/jamanetworkopen.2026.4003}
}

@article{kruse2025large,
  title   = {Large Language Models with Temporal Reasoning for Longitudinal Clinical Summarization and Prediction},
  author  = {Kruse, Maya and Hu, Shiyue and Derby, Nicholas and Wu, Yifu and Stonbraker, Samantha and Yao, Bingsheng and Wang, Dakuo and Goldberg, Elizabeth M. and Gao, Yanjun},
  journal = {Findings of the Association for Computational Linguistics: EMNLP 2025},
  pages   = {20715--20735},
  year    = {2025},
  doi     = {10.18653/v1/2025.findings-emnlp.1128}
}

@article{sheppert2026anchoring,
  title   = {Large language models exhibit greater diagnostic anchoring than physicians in a forced-choice vignette study},
  author  = {Sheppert, Alexander and Shen, Christine and Geissal, Erik and Riley, Craig and Riley, Sylvia},
  journal = {International Journal of Medical Informatics},
  volume  = {219},
  pages   = {106550},
  year    = {2026},
  doi     = {10.1016/j.ijmedinf.2026.106550}
}

@article{lapp2023dynamic,
  title   = {Dynamic Prediction of Patient Outcomes in the Intensive Care Unit: A Scoping Review of the State-of-the-Art},
  author  = {Lapp, Linda and Roper, Marc and Kavanagh, Kimberley and Bouamrane, Matt-Mouley and Schraag, Stefan},
  journal = {Journal of Intensive Care Medicine},
  volume  = {38},
  number  = {7},
  pages   = {575--591},
  year    = {2023},
  doi     = {10.1177/08850666231166349}
}

@article{li2026clicare,
  title   = {{CliCARE}: Grounding Large Language Models in Clinical Guidelines for Decision Support over Longitudinal Cancer Electronic Health Records},
  author  = {Li, Dongchen and Liang, Jitao and Li, Wei and Wang, Xiaoyu and Cao, Longbing and Yu, Kun},
  journal = {Proceedings of the AAAI Conference on Artificial Intelligence},
  volume  = {40},
  number  = {37},
  pages   = {31554--31562},
  year    = {2026},
  doi     = {10.1609/aaai.v40i37.40421}
}

@article{leibovitch2026prospective,
  title   = {Prospective evaluation of a large language model clinical decision support system in the emergency department},
  author  = {Leibovitch, Liron and Ahituv, Adi and Gorenshtein, Alon and Aran, Dvir and Sorka, Moran and Miron, Keren and others},
  journal = {Nature Medicine},
  year    = {2026},
  doi     = {10.1038/s41591-026-04601-5}
}

@article{agweyu2026generative,
  title   = {Generative AI-enabled clinical decision support system in primary care: a pragmatic, cluster-randomized trial},
  author  = {Agweyu, Ambrose and Mwaniki, Paul and Menon, Vaishnavi and Korom, Robert and Isaaka, Lynda and Wanyama, Conrad and Gill, Jaspret and Kiptinness, Sarah and Adan, Najib and Emmanuel-Fabula, Mira and Riley, Richard D. and Archer, Lucinda and Denniston, Alastair K. and Liu, Xiaoxuan and Mateen, Bilal A.},
  journal = {Nature Medicine},
  volume  = {32},
  pages   = {3032--3039},
  year    = {2026},
  doi     = {10.1038/s41591-026-04503-6}
}

@article{thirunavukarasu2023large,
  title   = {Large language models in medicine},
  author  = {Thirunavukarasu, Arun James and Ting, Darren Shu Jeng and Elangovan, Kabilan and Gutierrez, Laura and Tan, Ting Fang and others},
  journal = {Nature Medicine},
  volume  = {29},
  pages   = {1930--1940},
  year    = {2023},
  doi     = {10.1038/s41591-023-02448-8}
}

@article{vanveen2024adapted,
  title   = {Adapted large language models can outperform medical experts in clinical text summarization},
  author  = {Van Veen, Dave and Van Uden, Cara and Blankemeier, Louis and Delbrouck, Jean-Benoit and Aali, Asad and Bluethgen, Christian and Pareek, Anuj and Polacin, Malgorzata and Reis, Eduardo Pontes and others},
  journal = {Nature Medicine},
  volume  = {30},
  pages   = {1134--1142},
  year    = {2024},
  doi     = {10.1038/s41591-024-02855-5}
}

@article{johri2025evaluation,
  title   = {An evaluation framework for clinical use of large language models in patient interaction tasks},
  author  = {Johri, Shreya and Jeong, Jaehwan and Tran, Benjamin A. and Schlessinger, Daniel I. and Wongvibulsin, Shannon and Barnes, Leandra A. and Zhou, Hong-Yu and Cai, Zhuo Ran and Van Allen, Eliezer M. and Kim, David and Daneshjou, Roxana and Rajpurkar, Pranav},
  journal = {Nature Medicine},
  volume  = {31},
  pages   = {77--86},
  year    = {2025},
  doi     = {10.1038/s41591-024-03328-5}
}

@article{fleming2024medalign,
  author  = {Fleming, Scott L. and Lozano, Alejandro and Haberkorn, William J. and Jindal, Jenelle A. and Reis, Eduardo and Thapa, Rahul and Morse, Keith E. and Brunskill, Emma P. and Fries, Jason A. and Shah, Nigam H.},
  title   = {{MedAlign}: A Clinician-Generated Dataset for Instruction Following with Electronic Medical Records},
  journal = {Proceedings of the AAAI Conference on Artificial Intelligence},
  year    = {2024},
  volume  = {38},
  number  = {20},
  pages   = {22021--22030},
  doi     = {10.1609/aaai.v38i20.30205}
}

@article{cui2025timer,
  title   = {{TIMER}: Temporal Instruction Modeling and Evaluation for Longitudinal Clinical Records},
  author  = {Cui, Hejie and Unell, Alyssa and Chen, Bowen and Fries, Jason Alan and Alsentzer, Emily and Koyejo, Sanmi and others},
  journal = {npj Digital Medicine},
  volume  = {8},
  pages   = {577},
  year    = {2025},
  doi     = {10.1038/s41746-025-01965-9}
}

@article{tan2024generatedcontext,
  author  = {Tan, Hexiang and Sun, Fei and Yang, Wanli and Wang, Yuanzhuo and Cao, Qi and Cheng, Xueqi},
  title   = {Blinded by Generated Contexts: How Language Models Merge Generated and Retrieved Contexts When Knowledge Conflicts?},
  journal = {Proceedings of the 62nd Annual Meeting of the Association for Computational Linguistics},
  year    = {2024},
  pages   = {6207--6227},
  doi     = {10.18653/v1/2024.acl-long.337}
}

@article{kwan2024mteval,
  author  = {Kwan, Wai-Chung and Zeng, Xingshan and Jiang, Yuxin and Wang, Yufei and Li, Liangyou and Shang, Lifeng and Jiang, Xin and Liu, Qun and Wong, Kam-Fai},
  title   = {{MT}-Eval: A Multi-Turn Capabilities Evaluation Benchmark for Large Language Models},
  journal = {Proceedings of the 2024 Conference on Empirical Methods in Natural Language Processing},
  year    = {2024},
  pages   = {20153--20177},
  doi     = {10.18653/v1/2024.emnlp-main.1124}
}

@article{kamoi2024selfcorrection,
  title   = {When Can {LLM}s Actually Correct Their Own Mistakes? A Critical Survey of Self-Correction of {LLM}s},
  author  = {Kamoi, Ryo and Zhang, Yusen and Zhang, Nan and Han, Jiawei and Zhang, Rui},
  journal = {Transactions of the Association for Computational Linguistics},
  volume  = {12},
  pages   = {1417--1440},
  year    = {2024},
  doi     = {10.1162/tacl_a_00713}
}

@article{sharma2023sycophancy,
  title   = {Towards Understanding Sycophancy in Language Models},
  author  = {Sharma, Mrinank and Tong, Meg and Korbak, Tomasz and Duvenaud, David and Askell, Amanda and Bowman, Samuel R. and Cheng, Newton and Durmus, Esin and Hatfield-Dodds, Zac and Johnston, Scott R. and Kravec, Shauna and Maxwell, Timothy and McCandlish, Sam and Ndousse, Kamal and Rausch, Oliver and Schiefer, Nicholas and Yan, Da and Zhang, Miranda and Perez, Ethan},
  journal = {arXiv preprint arXiv:2310.13548},
  year    = {2023},
  doi     = {10.48550/arXiv.2310.13548}
}

@article{dvijotham2023deferral,
  title   = {Enhancing the reliability and accuracy of AI-enabled diagnosis via complementarity-driven deferral to clinicians},
  author  = {Dvijotham, Krishnamurthy and Winkens, Jim and Barsbey, Melih and Ghaisas, Sumedh and Stanforth, Robert and Pawlowski, Nick and others},
  journal = {Nature Medicine},
  year    = {2023},
  doi     = {10.1038/s41591-023-02437-x}
}

@article{pan2024automatically,
  author  = {Pan, Liangming and Saxon, Michael and Xu, Wenda and Nathani, Deepak and Wang, Xinyi and Wang, William Yang},
  title   = {Automatically Correcting Large Language Models: Surveying the Landscape of Diverse Automated Correction Strategies},
  journal = {Transactions of the Association for Computational Linguistics},
  volume  = {12},
  pages   = {484--506},
  year    = {2024},
  doi     = {10.1162/tacl_a_00660}
}

@article{bull2020repeated,
  author  = {Bull, Lucy M. and Lunt, Mark and Martin, Glen P. and Hyrich, Kimme and Sergeant, Jamie C.},
  title   = {Harnessing repeated measurements of predictor variables for clinical risk prediction: a review of existing methods},
  journal = {Diagnostic and Prognostic Research},
  volume  = {4},
  pages   = {9},
  year    = {2020},
  doi     = {10.1186/s41512-020-00078-z}
}

@article{andrinopoulou2021dynamic,
  author  = {Andrinopoulou, Eleni-Rosalina and Harhay, Michael O. and Ratcliffe, Sarah J. and Rizopoulos, Dimitris},
  title   = {Reflection on modern methods: Dynamic prediction using joint models of longitudinal and time-to-event data},
  journal = {International Journal of Epidemiology},
  volume  = {50},
  number  = {5},
  pages   = {1731--1743},
  year    = {2021},
  doi     = {10.1093/ije/dyab047}
}

\end{document}